\documentclass{article}

 \usepackage[main,final]{neurips_2026}

\usepackage[utf8]{inputenc} 
\usepackage[T1]{fontenc}    
\usepackage[colorlinks]{hyperref}
\usepackage{url}            
\usepackage{booktabs}       
\usepackage{amsfonts}       
\usepackage{nicefrac}       
\usepackage{microtype}      
\usepackage[dvipsnames,table]{xcolor}         
\usepackage{xspace}

\usepackage{xspace}
\newcommand*{\eg}{\emph{e.g.},\@\xspace}
\newcommand*{\ie}{\emph{i.e.},\@\xspace}
\newcommand*{\etc}{\emph{etc.}\@\xspace}

\definecolor{IllinoisOrange}{HTML}{FF5F05}
\definecolor{IllinoisBlue}{HTML}{13294B}

\usepackage{amsmath}
\usepackage{amssymb}
\usepackage{mathtools}
\usepackage{amsthm}
\usepackage{cleveref}
\usepackage{wrapfig}
\usepackage{float}
\usepackage{gradient-text}  
\usepackage{multirow}
\usepackage{enumitem}
\usepackage{pifont}
\usepackage{graphicx}
\usepackage{svg}

\usepackage{pgfplots}
\pgfplotsset{compat=1.18}
\usetikzlibrary{plotmarks}
\usepgfplotslibrary{external}

\newcommand{\cmark}{\textcolor{ForestGreen}{\ding{51}}}  
\newcommand{\xmark}{\textcolor{red}{\ding{55}}}

\definecolor{SoftBeige}{HTML}{FEF5E7}
\definecolor{YaleBlue}{HTML}{2A5487}
\definecolor{CustomRed}{RGB}{233, 93, 34}      
\definecolor{CustomOrange}{RGB}{245, 132, 38}  
\definecolor{CustomPeach}{RGB}{255, 165, 82}   
\definecolor{CustomLightOrange}{RGB}{255, 170, 120} 
\definecolor{CustomLightRedOrange}{RGB}{255, 195, 160} 
\definecolor{CustomYellow}{RGB}{255, 220, 148} 
\definecolor{CustomDarkGray}{RGB}{80, 80, 80}
\definecolor{CustomTeal}{HTML}{008080}   
\definecolor{BrightTeal}{HTML}{00A6A6}
\definecolor{CustomMint}{HTML}{50C99A}
\definecolor{CustomPink}{HTML}{FF828F}
\definecolor{CustomPurple}{HTML}{884AB2}
\definecolor{CustomLightPurple}{HTML}{C497FF}
\definecolor{CustomLightLightPurple}{HTML}{E6DEFF}
\definecolor{MyPurple}{RGB}{135,58,208}

 \usepackage[dvipsnames]{xcolor}
\newcommand{\modelname}{ELSA3D\xspace}
\newcommand{\modelnamenc}{ELSA3D\xspace}
\newcommand{\modelnamegradient}
{\textbf{\gradientRGB{ELSA3D}{255, 95, 5}{255, 0, 0}}\xspace}

\newcommand{\best}[1]{{\setlength{\fboxsep}{2pt}\colorbox{CustomPink}{\textbf{#1}}}}
\newcommand{\second}[1]{{\setlength{\fboxsep}{2pt}\colorbox{CustomLightRedOrange}{\underline{#1}}}}

\title{ELSA3D: Elastic Semantic Anchoring for Unified 3D Understanding and Generation}

\author{%
  \textbf{Tianjiao Yu, Xinzhuo Li, Yifan Shen, Onkar Susladkar,}\\\textbf{Yuanzhe Liu, Xiaona Zhou, Ismini Lourentzou} \\
  University of Illinois Urbana-Champaign\\
  \texttt{\{ty41,lourent2\}@illinois.edu} \\[5pt]
  \href{https://plan-lab.github.io/elsa3d}{\texttt{\textcolor{IllinoisOrange}{plan-lab.github.io/elsa3d}}}
}

\begin{document}

\maketitle

\begin{abstract}
 Unified 3D foundation models aspire to generate 3D assets and reason about them in language within a single backbone, but their text–3D interaction remains largely implicit. Existing methods concatenate text and 3D tokens into a flat sequence and rely on self-attention, collapsing coarse structural cues and fine geometric details into one undifferentiated representation. 
 We introduce \modelnamegradient, a unified 3D model that addresses this with \emph{elastic semantic anchoring}, structuring language and geometric reasoning jointly along matched abstraction scales. 
  \modelname{} represents geometry with a scale-aware octree tokenizer and introduces \emph{Anchor Tokens}, sparse cross-modal units that select semantic cues, route them to the most relevant 3D scale, retrieve scale-specific geometric evidence, and write the fused signal back into the unified representation,  keeping interaction sparse yet precise.
 A lightweight per-block router makes both computation and reasoning elastic, choosing which text tokens instantiate anchors at which geometric scale so that cross-modal capacity concentrates where alignment is most needed. 
 \modelname{} achieves state-of-the-art performance across image-to-3D generation, text-to-3D generation, and 3D captioning, outperforming the strongest unified baseline while roughly \emph{halving} FLOPs and inference latency relative to the non-elastic version of the same model.
\end{abstract}

\section{Introduction}
Unified 3D models~\cite{ye2025shapellm_omni, yu2025core3d, xu2026uniugg} aim to bridge 3D understanding and generation within one backbone, where a single model can reconstruct a 3D object from an image, generate one from language, describe its structure in text, and support downstream reasoning over geometry.
This unification is appealing because generation and understanding can support each other within a shared representation, but it also imposes stronger requirements on multimodal reasoning: a unified 3D model should balance global structure and local detail, translate open-ended language into concrete geometric decisions, and allocate compute dynamically when semantic-geometric alignment requires finer reasoning.

Current systems fall short on each of these requirements because text-3D interaction remains largely implicit. Previous works~\cite{ye2025shapellm_omni, yu2025core3d} concatenate text and 3D tokens into a monolithic sequence and rely on self-attention to discover cross-modal correspondences. Recent 3D advances~\cite{dutt2026lost, zhang2025vertexregen, chen2025sar3d, chen2023neural} make the geometric representation multiscale, but the reasoning architecture has not evolved accordingly. What is missing, therefore, is not merely a stronger hierarchical 3D representation, but a unified design that structures language reasoning and geometric reasoning jointly, enabling \emph{structured interaction} between semantic cues and geometric content.

To address this gap, we introduce \modelnamegradient{}, a unified 3D model built around \emph{elastic semantic anchoring}. \modelname{} first represents 3D shapes with an octree VQ-VAE in which every content token carries an explicit deterministic scale tag, exposing multiple geometric resolutions to the model. Then, the model organizes language into a semantic trace spanning \emph{Global}, \emph{Structure}, and \emph{Appearance} cues, decomposing text descriptions into finer semantic granularity.
To connect both semantic and geometric abstractions, \modelname{} introduces \emph{Anchor Tokens}. Each anchor is a transient cross-modal unit instantiated from a selected semantic token, routed to the most relevant 3D scale, fused with scale-specific geometric evidence, and written back into the unified sequence. Anchors keep cross-modal interaction sparse and explicit at once, avoiding the cost of dense text-geometry attention while preserving precise binding where it matters.

Since not all semantic tokens contribute equally to 3D grounding (\eg function words, generic modifiers, \etc), \modelname{} equips each transformer block with a lightweight \emph{elastic router}. The router selects which text tokens instantiate anchors and which 3D scale they query, while also deciding whether the block should execute and how much MLP width to allocate. This makes both computation and reasoning elastic by focusing on cross-modal interaction where targeted reasoning is most needed.  

Extensive experiments demonstrate that elastic semantic anchoring improves unified 3D models along generation fidelity, 3D-language understanding, and inference efficiency.
\modelname{} achieves state-of-the-art results across image-to-3D generation, text-to-3D generation, and 3D captioning, improving over the strongest unified baseline by $+2.74$ CLIP and $-2.03$ FD on image-to-3D, $+1.35$ CLIP and $-4.05$ FD on text-to-3D, and $+3.56$ Sentence-BERT on 3D captioning. 
Ablations further show that sparse anchor routing outperforms dense cross-modal fusion while reducing FLOPs from 1081G to 632G and latency from 29.8s to 17.2s.

Our contributions are summarized as follows:
\begin{itemize}[itemsep=0.5ex, parsep=0pt, topsep=-2.3pt, leftmargin=0.4cm]
    \item We introduce \modelnamegradient{}, a unified 3D understanding-and-generation model that structures language reasoning and geometric reasoning along matched abstraction scales.

    \item We propose \emph{Anchor Tokens}, a sparse and dynamic cross-modal interface that grounds selected semantic tokens in scale-specific 3D evidence and writes the fused language-geometry signal back into the unified representation.

    \item We design an elastic routing mechanism that jointly controls block execution, MLP width, anchor selection, and geometric scale assignment, enabling elastic computation and elastic semantic-geometric reasoning under a single routing scheme.

    \item We demonstrate that scale-aware semantic anchoring improves 3D generation and captioning, while joint depth--width adaptation preserves near-full-compute quality at roughly half the FLOPs and inference latency. Ablations isolate the benefits of semantic decomposition, sparse anchors, and learned scale routing.
\end{itemize}

\section{Related Work}
Our work is closely related to {unified 3D understanding and generation models}~\cite{ye2025shapellm_omni,yu2025core3d,xu2026uniugg}, {hierarchical 3D representation and generation}~{\cite{liu2024texoct,dutt2026lost,zhang2025vertexregen}}, {adaptive token selection and elastic computation}~\cite{bolyatoken,liang2022not,zhao2024dynamic,zhu2025ea,wu2023elastic}, and {multimodal token fusion and alignment}~\cite{wang2022multimodal,zhang2024magic,zhong2025aim,li2025flowmm,xu2022groupvit,yin2025sea,zhang2025attanchor}. A comprehensive discussion is provided in Appendix~\ref{app:related}. 
Prior hierarchical 3D models demonstrate that multiscale structure is effective for scalable representation and generation, but they use hierarchy primarily to organize geometry, whereas prior routing and token-selection methods mainly use routing to reduce computation, prune redundant tokens, or allocate tokens across experts. \modelname{} instead couples hierarchy and routing for scale-aware language--geometry grounding: the hierarchy provides explicit geometric levels of detail, and routing determines which scale each semantic anchor should query before writing fused cross-modal evidence back into the unified understanding-and-generation model.\looseness-1

\section{Method}
\label{sec:method_main}

\begin{figure*}[t]
    \centering
    \includegraphics[width=0.99 \textwidth]{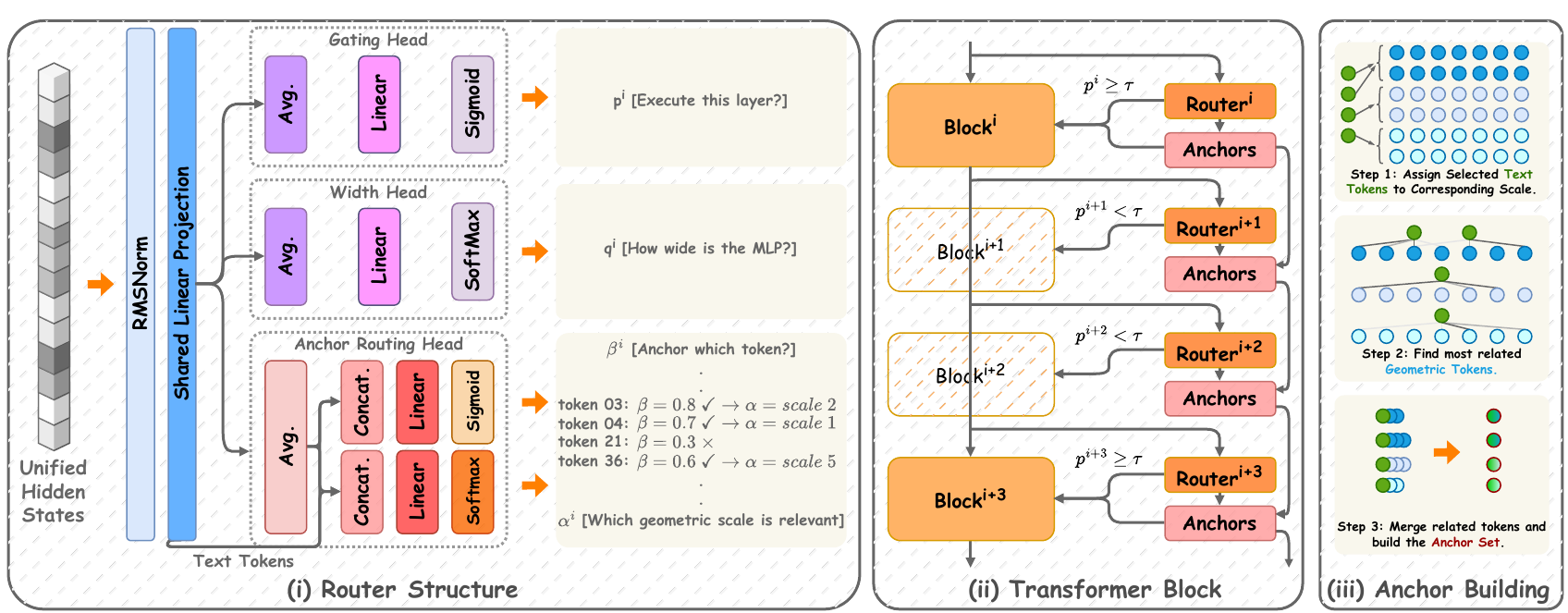}
    \vspace{-0.2cm}
    \caption{\textbf{\modelname{} overview.}
    \modelname{} is built around \emph{elastic semantic anchoring}, where routing jointly controls computation and semantic--geometric grounding. (i) The router has three heads: a Gating Head ($p^i$, skip or run), a Width Head ($q^i$, MLP width), and an Anchor Routing Head ($\beta^i, \alpha^i$, which text tokens become anchors and at which scale). (ii) Blocks with $p^i \geq \tau$ execute at the selected width; others are skipped. (iii) Selected text tokens are routed to their preferred scale, cross-attended to the 3D tokens at that scale, and fused into the anchor set.}
    \label{fig:main_method}
    \vspace{-0.1cm}
\end{figure*}

A core challenge in unified 3D foundation models is translating open-ended language into structured reasoning signals that preserve compositional semantics and physical constraints~\cite{ye2025shapellm_omni,xu2026uniugg,yu2025core3d}. Directly mapping text prompts to latent 3D tokens is inherently under-specified: language descriptions typically omit precise geometric details, relations, and material cues, yielding shapes that match coarse appearance but break in structure and texture consistency. Yet exhaustively coupling every text token to all 3D tokens is both computationally wasteful and semantically noisy, since many tokens lack precise geometric counterparts.
\modelname{} addresses this mismatch with {elastic semantic anchoring} (\Cref{fig:main_method}). First, we represent each 3D object with a multiscale octree VQ-VAE whose tokens carry explicit \emph{scale tags}, making geometric resolution available to the transformer (\Cref{sec:representations}). Second, we introduce \emph{anchor tokens}, transient cross-modal units that bind selected semantic tokens to geometric evidence at a specific scale and write the fused signal back into the unified sequence (\Cref{sec:alignment}). Third, we use a lightweight per-block router to make both computation and grounding adaptive (\Cref{sec:design}). \modelname{} is trained in two stages: the octree VQ-VAE is first trained on 3D data, then the unified LLM {is trained} over interleaved text and 3D sequences with auxiliary budget losses that shape the router's depth, width, and anchor-sparsity decisions (\Cref{sec:training}).

\subsection{Semantic and Geometric Representations}
\label{sec:representations}

Given an input condition $x$ (\eg a text prompt, an image, or both), \modelname{} generates an output sequence $y$ that represents either a 3D object or a language response grounded in 3D geometry. For 3D generation, $y$ consists of structural and content tokens that define an octree-based 3D representation, which is decoded into a textured 3D shape. For 3D understanding, $y$ is a language sequence conditioned on 3D geometry. We model both tasks with a unified autoregressive transformer over semantic tokens and geometric tokens.
At transformer block $i$, we denote the hidden states of the unified sequence by
$\mathbf{H}^{i}=\{\mathbf{h}^{i}_{j}\}_{j=1}^{N_{\mathrm{uni}}}$,
where $N_{\mathrm{uni}}$ is the total sequence length. The semantic-token subset is
$\mathbf{T}^{i}=\{\mathbf{t}^{i}_{m}\}_{m=1}^{M}$.
The geometric tokens are organized by octree scale, where $\mathbf{G}^{i,s}$ denotes the hidden states of geometric tokens at scale $s$, with $s\!=\!1$ being the coarsest scale and $s\!=\!S$ the finest scale.
This unified formulation requires representations that make both semantic abstraction and geometric resolution explicit. We therefore first define the semantic and geometric traces used by the model, then describe how anchor tokens dynamically couple them.

\textbf{Semantic Representation.}
Language specifies a 3D object through cues at different abstraction levels, including category and global shape, part structure and proportions, and surface appearance. To expose this structure to the model, we organize semantic tokens into a {semantic trace} with three aspects. The \emph{Global} aspect captures the object's category, overall silhouette, and dominant orientation, the \emph{Structure} aspect describes mass distribution, proportions, and part composition, and the \emph{Appearance} aspect captures material, color, texture, and local detail. This trace acts as an interpretable, language-level scaffold that anchors the subsequent geometric reasoning.

\begin{figure*}[t]
    \centering
    \includegraphics[width=0.99\textwidth]{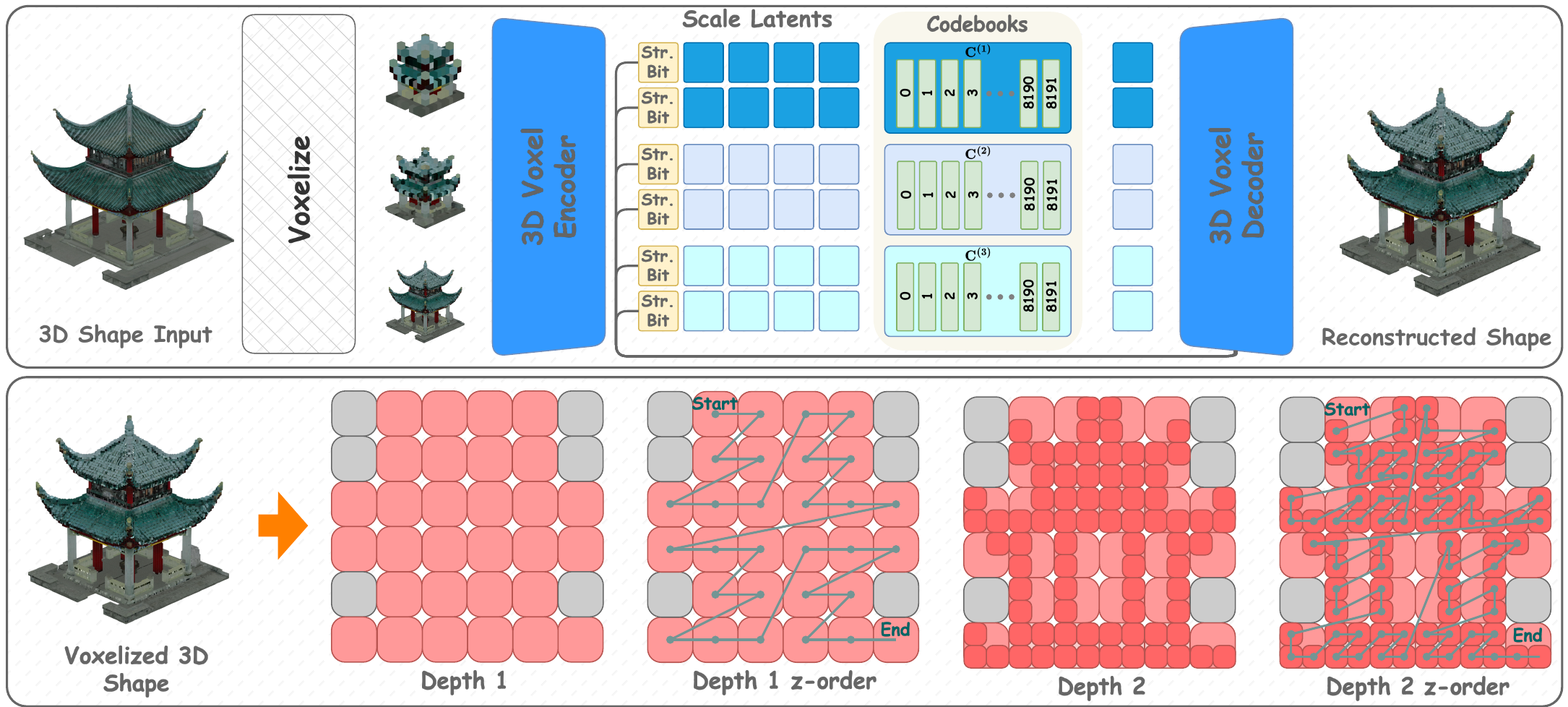}
    \vspace{-0.2cm}
    \caption{\textbf{Scale-aware octree tokenization.} Top: \modelname{}'s octree VQ-VAE encodes a voxelized 3D shape into multiscale structural bits and scale-specific content codes, then decodes them to reconstruct the shape. Bottom: nodes are organized by octree depth and serialized with Morton/Z-order to preserve spatial locality within each scale.}
    \label{fig:octree_representation}
    \vspace{-0.1cm}
\end{figure*}

\textbf{Geometric Representation.}
We represent each 3D object as a multiscale octree constructed from a canonicalized $128^{3}$ voxel grid. Starting from the unit cube, space is recursively subdivided to a maximum depth $S$. To provide a stable global scaffold, we fully populate the first three octree levels: every node at these depths is present regardless of surface occupancy. Beyond this coarse scaffold, the octree becomes sparse, and nodes are added only in regions that intersect surface geometry. 
As illustrated in \Cref{fig:octree_representation}, within each depth, nodes are serialized in Morton (Z-order), preserving spatial locality in the token sequence~{\cite{wang2023octformer}}.
Formally, each octree node $v$ is represented by a {structural bit} $o_v \in \{0,1\}$ indicating whether $v$ is subdivided, and a {content token} $\mathbf{g}_v$ encoding its local geometry. Structure bits are used to reconstruct the octree topology during decoding. We encode node geometry using scale-specific vector quantization. Inspired by previous octree-based works~\cite{wang2023octformer, wei2025octgpt}, for each node $v$ at scale $s$, we extract a feature vector $\mathbf{f}_v \in \mathbb{R}^{C}$ and quantize it against a scale-specific codebook $\mathcal{C}^{(s)}$. The discrete code index and corresponding codebook embedding are
\begin{equation}
\setlength{\abovedisplayskip}{3pt}
\setlength{\belowdisplayskip}{3pt}
\setlength{\abovedisplayshortskip}{2pt}
\setlength{\belowdisplayshortskip}{2pt}
{k_v = \arg\min_{k\in\{1,\ldots,|\mathcal{C}^{(s)}|\}} \|\mathbf{f}_v - \mathcal{C}^{(s)}[k]\|_2},
\qquad
\mathbf{g}_v = \mathcal{C}^{(s)}[k_v].
\end{equation}
Using separate codebooks per scale allows each vocabulary to specialize to geometric primitives at its resolution. However, quantization collapses geometrically similar regions onto the same codebook entry, erasing their spatial identity. 
We therefore augment each content token with a learned positional embedding $\mathbf{E}_{\text{pos}}^{(s)}[v]$ indexed by the node's Morton position at scale $s$, and a deterministic scale tag $\mathbf{s}^{\text{det}}_s$.
The scale tag is a fixed, non-trainable vector appended to each token, so a token's scale is recoverable from its embedding at initialization. This explicit scale signal is also needed by the alignment module (\Cref{sec:alignment}). The augmented content token is
\(\tilde{\mathbf{g}}_v\!=\![\,\mathbf{g}_v + \mathbf{E}_{\text{pos}}^{(s)}[v]\,;\,\mathbf{s}^{\text{det}}_s\,]. \)
We denote by $\mathbf{G}^{s}$ the Morton-ordered sequence of augmented structural-content pairs $(o_v,\tilde{\mathbf{g}}_v)$ at scale $s$, and define the full geometric trace as
\(
\mathcal{G}_{\text{geo}}\!=\![\mathbf{G}^{1}; \mathbf{G}^{2}; \dots; \mathbf{G}^{S}].
\)
At transformer block $i$, we write $\mathbf{G}^{i,s}$ for the hidden states corresponding to the geometric tokens at scale $s$, initialized from $\mathbf{G}^{s}$ at the input layer. While the semantic trace provides the conceptual plan in language space, this representation organizes the geometric trace into a natural coarse-to-fine hierarchy with an explicit scale signal on every token.

\subsection{Anchor Tokens}
\label{sec:alignment}
Dense interaction between all semantic tokens and all 3D tokens is costly and often unnecessary, as many words provide global or contextual constraints, while only a subset requires precise geometric grounding. To improve reasoning between language and 3D, we therefore introduce \textbf{\emph{Anchor Tokens}}, a sparse semantic-geometric interface that creates cross-modal interaction only where it is useful. An anchor is a block-local fusion unit formed from a selected semantic token and the geometric evidence it retrieves from a routed 3D scale. Unlike fixed alignments or dense text--3D attention, anchors are constructed dynamically for each input and transformer block, allowing the model to allocate cross-modal capacity to the semantic cues that most require geometric evidence.

\textbf{Anchor Construction.}
At block $i$, the router introduced in \Cref{sec:design} emits anchor-routing signals $(\beta_m^i, \boldsymbol{\pi}_m^i, \alpha_m^i)$ for each semantic token $\mathbf{t}_m^i$. Here, $\beta_m^i\in[0,1]$ is an anchor gate that indicates whether the $m$-th text token should participate in anchor construction, $\boldsymbol{\pi}_m^i\in\Delta^{S-1}$ is a distribution over geometric scales, and
\(
\alpha_m^i = \arg\max_{s\in\{1,\dots,S\}} \pi_{m,s}^i
\)
is the selected geometric scale.
Let $\mathcal{M}^i=\{m:\beta_m^i\ge\tau_a\}$ denote the selected semantic-token indices at block $i$.
For each selected token $m\in\mathcal{M}^i$, we gather scale-specific geometric evidence by cross-attention:
\begin{equation}
\setlength{\abovedisplayskip}{3pt}
\setlength{\belowdisplayskip}{3pt}
\setlength{\abovedisplayshortskip}{2pt}
\setlength{\belowdisplayshortskip}{2pt}
\mathbf{e}_m^{i} = 
\mathrm{CrossAttn}\!\left(\mathbf{t}_m^{i},\, \mathbf{G}^{i,\alpha_m^i}\right),
\end{equation}
where the text token $\mathbf{t}_m^{i}$ serves as the query and the 3D tokens at the selected scale $\alpha_m^i$ serve as keys and values.  The anchor token is then formed by fusing the semantic token with the aligned geometric evidence:
\(
\mathbf{a}_m^{i} =
\mathrm{MLP}_{\mathrm{anchor}}\!\left([\mathbf{t}_m^{i};\,\mathbf{e}_m^{i}]\right).
\)
Each anchor is therefore a compact semantic-geometric interaction unit grounded in both the current text state and a specific level of 3D context. The number of anchor tokens is dynamic: at block $i$, only the subset of text tokens selected by $\beta_m^i$ instantiates anchors, so $|\mathbf{A}^i|$ varies across layers and inputs.

\textbf{Write-back.}
The resulting anchor set $\mathbf{A}^{i} = \{\mathbf{a}_m^i \mid m\in\mathcal{M}^i\}$ acts as a transient block-local cross-modal workspace. To make anchors influence the persistent unified sequence, every token in the main sequence cross-attends to the current anchor set:
\begin{equation}
\mathbf{d}_j^{i} = \mathrm{CrossAttn}_{\mathrm{wb}}\!\big(\mathbf{h}_j^{i},\, \mathbf{A}^{i}\big), \qquad j = 1,\dots,N_{\mathrm{uni}},
\end{equation}
where $\mathbf{h}_j^{i}$ is the $j$-th hidden state of the persistent sequence entering block~$i$. This lets both text and 3D tokens absorb the fused semantic-geometric signal. A learned per-token gate controls the injection magnitude, preventing anchors from dominating early in training:
\begin{equation}
\gamma_j^{i} = \sigma\!\big(\mathrm{Linear}([\mathbf{h}_j^{i};\, \mathbf{d}_j^{i}])\big), \qquad
\mathbf{h}_j^{i,\text{+a}} = \mathbf{h}_j^{i} + \gamma_j^{i} \odot \mathbf{d}_j^{i}.
\end{equation}
The gate bias is initialized to $-2$ so that $\gamma_j^{i}\!\approx\!0$ at the start of training, letting anchors gradually earn influence. The anchor-augmented sequence $\mathbf{H}^{i,\text{+a}}$ then replaces $\mathbf{H}^{i}$ as the input to block~$i$'s self-attention and feed-forward sub-layers (see~\Cref{sec:design}). After the block update, anchors are discarded and only the persistent sequence continues to the next layer.
If no anchors are selected, the write-back is skipped and $\mathbf{H}^{i,\mathrm{+a}}=\mathbf{H}^{i}$. 
Since $|\mathbf{A}^i|\!\ll\!N_{\mathrm{uni}}$, the write-back cross-attention adds only $O(N_{\mathrm{uni}}\!\cdot\!|\mathbf{A}^i|)$ cost per block, which is negligible compared to the main self-attention.

\subsection{Dynamic Routing and Elastic Reasoning}
\label{sec:design}
Existing unified 3D models~\cite{ye2025shapellm_omni, xu2026uniugg,li2024uni3dl, yu2025core3d} typically use a fixed transformer computation pattern and rely on implicit token mixing for cross-modal exchange. This treats all blocks as equally necessary and leaves semantic-geometric grounding to emerge without explicitly deciding which language tokens should query which geometric scale. In practice, both assumptions are limiting: input difficulty varies across examples, and only a subset of semantic tokens requires precise geometric grounding~\cite{liang2022not,ye2026not}. We therefore introduce a lightweight router that makes reasoning \emph{elastic} in both computation and grounding. Inspired by dynamic transformer designs~\cite{zhu2025ea, zhao2024dynamic, wang2026elastic}, the router adaptively controls computation by deciding block execution and MLP width. More importantly, it also controls the cross-modal communication pattern by selecting which semantic tokens instantiate anchor tokens and at which geometric scale each should ground.

\textbf{Router Architecture.}
Given a unified transformer backbone with $n$ blocks $\{\mathbf{B}^{i}\}_{i=1}^{n}$, we attach a lightweight three-headed router $\mathbf{R}^{i}$ to each block. Let $\mathbf{H}^{i}\in\mathbb{R}^{N_{\mathrm{uni}}\times D}$ denote the hidden states of the persistent unified sequence entering block $i$, where $N_{\mathrm{uni}}$ is the total number of semantic and geometric tokens, and $D$ is the hidden dimension. Anchor tokens are auxiliary block-local units and are not included in this persistent sequence length. The router first projects each token into a low-dimensional routing space
\(
\mathbf{r}_l^{i} = \mathbf{h}_l^{i}\mathbf{W}_r, 
~~l = 1,\dots,N_{\mathrm{uni}},
\)
where $\mathbf{W}_r \in \mathbb{R}^{D\times d_r}$ and $\mathbf{r}_l^i\in\mathbb{R}^{d_r}$. These per-token routing features are mean-pooled to obtain a block-level context
\(
\setlength{\abovedisplayskip}{3pt}
\setlength{\belowdisplayskip}{3pt}
\setlength{\abovedisplayshortskip}{2pt}
\setlength{\belowdisplayshortskip}{2pt}
\mathbf{b}^{i} = \frac{1}{N_{\mathrm{uni}}}\sum_{l=1}^{N_{\mathrm{uni}}}\mathbf{r}_l^{i}.
\)
The first two router heads use $\mathbf{b}^i$ for block-level computation decisions: (1) a scalar \emph{block-gating logit} $\ell^{i}=\mathbf{b}^{i}\mathbf{w}_{\ell}$, which determines whether block $i$ is executed, and (2) a \emph{width logit vector} $\mathbf{u}^{i}=\mathbf{b}^{i}\mathbf{W}_{u}$, which determines the MLP width used by the block.
The third head operates at the token level over the semantic-token subset. For each semantic token $\mathbf{t}_m^i$, it uses the token routing feature $\mathbf{r}_m^i$ together with the block context $\mathbf{b}^i$ to predict \emph{anchor-routing signals} $(\beta_m^i,\boldsymbol{\pi}_m^i)$, where $\beta_m^i$ gates whether semantic token $m$ instantiates an anchor, and $\boldsymbol{\pi}_m^i$ is a distribution over geometric scales.

\textbf{Adaptive Block Skipping.}
From the block-gating logit, we compute an execution probability
\begin{equation}
    p^{i} = \sigma(\ell^{i}) \in [0,1].
\end{equation}
Block $\mathbf{B}^i$ is executed when $p^i\ge\tau$ and skipped otherwise; we use $\tau=0.5$ throughout. When a block is skipped, anchor construction and write-back are also skipped, so $\mathbf{H}^{i,+a}=\mathbf{H}^{i}$.  Each block residual update is modulated by gate $\delta^i=\mathbf{1}[p^i\ge\tau]$, yielding  
\begin{equation}
\mathbf{H}^{i+1}
=
\mathbf{H}^{i,+a}
+
\delta^{i}
\big(
\mathbf{B}^{i}(\mathbf{H}^{i,+a})-\mathbf{H}^{i,+a}
\big).
\end{equation}
Because $\delta^{i}$ is discrete, we use a {Straight-Through Estimator}~\cite{bengio2013ste} during training and replace $\delta^{i}$ with $\mathbf{1}[p^{i}\ge\tau] + p^{i} - \mathrm{sg}(p^{i})$ in the backward pass, where $\mathrm{sg}(\cdot)$ denotes stop-gradient.

\textbf{Adaptive MLP Width.}
Skipping a block is a coarse decision. For executed blocks, the router additionally selects the feed-forward capacity. 
Let $H$ be the full hidden width of the MLP. We discretize the width choice into four levels $\mathcal{W} = \{\tfrac{1}{4},\tfrac{1}{2},\tfrac{3}{4},1\}$, so the router can trade capacity for compute in meaningful increments. From the width logits $\mathbf{u}^{i}$, we obtain a distribution over levels and select the most probable one, \ie
\(
    \mathbf{q}^{i} = \mathrm{softmax}(\mathbf{u}^{i}),
    \qquad
    \hat{w}^{i} = \mathcal{W}\big[\arg\max_{j}\, q^{i}_{j}\big].
 \)
Such a hard weight slice would block gradients from flowing to the unused columns during training. We therefore apply a binary channel mask $\mathbf{m}(\hat{w}^{i})\!\in\!\{0,1\}^{H}$ that keeps the first $\hat{w}^{i}\!\cdot\!H$ intermediate channels:
\begin{equation}
    \mathrm{MLP}_{\text{adapt}}(\mathbf{z}) = \Big(\sigma(\mathbf{z}\mathbf{W}_{1})\odot \mathbf{m}(\hat{w}^{i})\Big)\,\mathbf{W}_{2}.
\end{equation}
This preserves a shared parameterization across all width choices during training, while producing the same forward computation as a narrowed MLP. 

\textbf{Anchor Routing Head.}
The anchor-routing head determines the structure of cross-modal grounding. Unlike the block-gating and width heads, which make block-level decisions from $\mathbf{b}^i$, anchor routing is token-specific. For each semantic token $\mathbf{t}_m^i$, we concatenate its routing feature with the block context:
\(
\mathbf{z}_m^i = [\mathbf{r}_m^i;\mathbf{b}^i],
\)
and predict an anchor gate and a soft distribution over 3D scales:
\begin{equation}
\beta_m^{i}=\sigma(\mathbf{z}_m^{i}\mathbf{W}_{\beta}),
\qquad
\boldsymbol{\pi}_m^{i}=\mathrm{softmax}(\mathbf{z}_m^{i}\mathbf{W}_{\pi}).
\end{equation}
Here $\beta_m^i\in[0,1]$ controls whether token $m$ instantiates an anchor, while $\boldsymbol{\pi}_m^i=[\pi_{m,1}^i,\dots,\pi_{m,S}^i]$ defines its preference over the $S$ geometric scales. The hard scale assignment is
\( 
\alpha_m^i\!=\!\arg\max_{s\in\{1,\dots,S\}}\pi_{m,s}^{i}.
\)
A semantic token is selected when $\beta_m^i\ge\tau_a$, where $\tau_a=0.5$ in all experiments. As with block skipping, $\arg\max$ is non-differentiable; we use a straight-through estimator so that gradients flow back through $\boldsymbol{\pi}_m^{i}$ during training.

For each selected semantic token, the anchor module in \Cref{sec:alignment} cross-attends to 3D tokens at scale $\alpha_m^i$, retrieves scale-specific geometric evidence, and fuses this evidence with the semantic state to form an anchor. 
By first selecting a geometric scale and then attending within that scale, this process induces a coarse-to-fine search, allowing each text token to quickly converge on its most relevant geometric correspondence.

\textbf{Training and Inference.}
\label{sec:training}
Training proceeds in two stages. We first train the scale-aware octree VQ-VAE on 3D data to obtain discrete multiscale geometry tokens. We then freeze the tokenizer and train the unified autoregressive model over interleaved semantic tokens, structural bits, and 3D content codes, together with auxiliary losses that shape the router's compute and grounding decisions.  
Additional details, including VQ-VAE training, unified autoregressive training, and router regularization, are provided in Appendix~\ref{app:training}.
\begin{table}[t]
  \setlength{\tabcolsep}{8pt}
  \centering
  \caption{\textbf{Image-to-3D Generation.}
  Methods are grouped by conditioning regime and unified modeling capability. 
  {\best{\textbf{Best}}} and 
  {\colorbox{CustomLightRedOrange}{{second-best}}} results are highlighted.}
  \label{tab:quant_image_to_3d}
  \vspace{-0.3cm}
  \resizebox{0.99\textwidth}{!}{%
  \begin{tabular}{l | ccccccc}
    \toprule
    \textbf{Model}
    & \textbf{CLIP} $\uparrow$ 
    & \textbf{FD} $\downarrow$ 
    & \textbf{KD} $\downarrow$ 
    & \textbf{PSNR} $\uparrow$ 
    & \textbf{LPIPS} $\downarrow$ 
    & \textbf{COV(\%)} $\uparrow$ 
    & \textbf{MMD(\textperthousand)} $\downarrow$ \\
    \midrule

    \rowcolor{gray!14}
    \multicolumn{8}{@{}l}{\textbf{\textit{Image-conditioned 3D generation}}} \\
    \midrule
    Direct3D 
    & 74.12 & 24.97 & 0.33 & 22.36 & 0.17 & 58.72 & 18.46 \\
    InstantMesh 
    & 84.41 & 20.13 & 0.29 & 25.72 & 0.11 & 66.84 & 16.72 \\

    \midrule
    \rowcolor{gray!14}
    \multicolumn{8}{@{}l}{\textbf{\textit{Text- and image-conditioned 3D generation}}} \\
    \midrule
    Shap-E 
    & 80.16 & 34.64 & 0.87 & 16.84 & 0.21 & 61.41 & 19.19 \\
    LN3Diff 
    & 82.79 & 26.98 & 0.76 & 18.73 & 0.19 & 55.21 & 19.84 \\
    XCube 
    & 84.91 & 10.32 & 0.09 & 23.99 & 0.13 & \second{{73.01}} & 14.92 \\
    SAR3D 
    & 84.67 & 22.12 & 0.18 & 26.31 & \second{{0.10}} & 70.30 & 15.12 \\
    3DTopia-XL 
    & 76.46 & 24.21 & 0.29 & 22.06 & 0.18 & 58.93 & 17.62 \\
    Gau.Any. 
    & 80.91 & 22.46 & 0.44 & 23.84 & 0.15 & 60.01 & 15.47 \\
    Trellis 
    & 85.03 
    &\second{{10.31}} 
    &\second{{0.08}} 
    & 24.01 
    & 0.14 
    & 72.10 
    & 14.36 \\

    \midrule
    \rowcolor{gray!14}
    \multicolumn{8}{@{}l}{\textbf{\textit{Unified 3D understanding and generation}}} \\
    \midrule
    Shap.-Omni 
    & 84.54 & 12.22 & 0.09 & 25.96 & 0.12 & 71.84 & 14.61 \\
    CoRe3D 
    & \second{{86.47}} 
    & 11.26 
    &\second{{0.08}} 
    & \second{{27.38}} 
    & 0.11 
    & 72.64 
    &\second{{14.28}} \\
    \rowcolor{gray!8} \textbf{\modelnamegradient} 
    & \best{\textbf{89.21}} 
    & \best{\textbf{9.23}} 
    & \best{\textbf{0.06}} 
    &\best{{\textbf{29.46}}} 
    &\best{{\textbf{0.06}}} 
    & \best{\textbf{75.84}} 
    & \best{\textbf{13.74}} \\
    \bottomrule
  \end{tabular}%
  }
  \vspace{-0.3cm}
\end{table}

\section{Experiments}
\label{sec:experiments}

We evaluate \modelnamenc{} across four capabilities: image-conditioned 3D generation, text-conditioned 3D generation, 3D object captioning, and general conversational reasoning.
For training, we use the publicly available 3D-Alpaca dataset~\cite{ye2025shapellm_omni} and additional 3D assets from Trellis-500K~\cite{xiang2025structured}, curated from ObjaverseXL~\cite{deitke2023objaverse}, ABO~\cite{Collins2022abo}, 3DFUTURE~\cite{fu20213d}, and HSSD~\cite{khanna2024habitat}. We also include UltraChat~\cite{ding2023ultrachat} to preserve general language capability.
For evaluation, we use Toys4K assets~\cite{stojanov2021using} and 200 in-the-wild images, with no overlap with the training set. Full implementation details are provided in Appendix~\ref{app:implement_details}.

\noindent\textbf{Baselines.}
{For 3D generation, we compare with representative open-source methods, including Direct3D~\cite{wu2024direct3d}, InstantMesh~\cite{xu2024instantmesh}, {Shap-E}~\cite{jun2023shap}, LN3Diff~\cite{lan2024ln3diff}, XCube~\cite{ren2024xcube}, SAR3D~\cite{chen2025sar3d}, 3DTopia-XL~\cite{chen20253dtopia}, GaussianAnything~\cite{yushi2025gaussiananything}, Trellis~\cite{xiang2025structured}, ShapeLLM-Omni~\cite{ye2025shapellm_omni}, and CoRe3D~\cite{yu2025core3d}.} For 3D object captioning, we compare with general VLMs LLaVA-13B~\cite{liu2023llava} and Qwen2.5-VL-7B~\cite{bai2025qwen25vl}, 3D-specialized understanding models 3D-LLM~\cite{hong20233d}, LEO~\cite{huang2023embodied}, and PointLLM-13B~\cite{xu2024pointllm}, and unified 3D generative models ShapeLLM-Omni~\cite{ye2025shapellm_omni} and CoRe3D~\cite{yu2025core3d}.

\noindent \textbf{Evaluation Metrics and Benchmarks.}
For 3D generation, we report CLIP similarity~\cite{radford2021learning} to measure consistency between the generated results and the input, FD~\cite{heusel2017gans} and KD~\cite{binkowski2018demystifying} (with Inception-v3~\cite{szegedy2015inception} as the feature extractor) to assess overall generation quality, PSNR and {LPIPS~\cite{zhang2018unreasonable}} to evaluate visual reconstruction quality, and COV and MMD~\cite{achlioptas2018learning} to measure distribution-level fidelity. We additionally use Q-Align~\cite{wu2023q}, a learned visual quality score that aligns with human perceptual judgments of generated 3D renderings.
For 3D captioning, we follow PointLLM~\cite{xu2024pointllm} and report 
BLEU-1~\cite{papineni2002bleu}, ROUGE-L~\cite{lin2004rouge}, METEOR~\cite{banerjee2005meteor}, Sentence-BERT~\cite{reimers2019sentence} and SimCSE~\cite{gao2021simcse}. 
For general language ability, we evaluate on MMLU~\cite{hendrycks2020measuring}, PIQA~\cite{bisk2020piqa}, GSM8K~\cite{cobbe2021training}, and SIQA~\cite{sap2019socialiqa}.

\noindent\textbf{Image-to-3D Generation.}
{\Cref{tab:quant_image_to_3d} shows that \modelnamenc{} achieves the best result on all seven metrics among the compared methods.}
Compared with the strongest unified baseline, our method improves by +2.74 CLIP, $-$2.03 FD, and $-$0.54 MMD, indicating that elastic semantic anchoring improves cross-modal grounding without sacrificing geometric quality.
For visual reconstruction quality, our method improves PSNR by 2.08 dB over CoRe3D (27.38 to 29.46) and reduces LPIPS by 40\% relative to SAR3D (0.10 to 0.06), the strongest compared baselines on the respective metrics.

\begin{wraptable}{r}{0.54\textwidth}
  \vspace{-0.5cm}
  \centering
  \renewcommand{\arraystretch}{1.08}
  \setlength{\tabcolsep}{4.0pt}
  \caption{\textbf{Text-to-3D Generation.} \modelnamegradient{} achieves {the} \best{best} performance across all metrics.}
  \label{tab:quant_text_to_3d}
  \resizebox{\linewidth}{!}{%
  \begin{tabular}{@{}lcccc@{}}
    \toprule
    \textbf{Method}
    & \textbf{CLIP} $\uparrow$
    & \textbf{FD} $\downarrow$
    & \textbf{KD} $\downarrow$
    & \textbf{Q-Align} $\uparrow$ \\
    \midrule

    \rowcolor{gray!14}
    \multicolumn{5}{@{}l}{\textbf{\textit{Text-/image-conditioned generators}}} \\
    Shap-E
    & 24.94 & 53.24 & 1.13 & 1.45 \\
    LN3Diff
    & 18.79 & 68.09 & 2.24 & 2.14 \\
    XCube
    & 26.37 & 31.82 & 0.42 & 1.68 \\
    SAR3D
    & 23.21 & 22.43 & 0.23 & 2.91 \\
    3DTopia-XL
    & 25.89 & 43.46 & 1.18 & 1.47 \\
    Gau.Any.
    & 24.76 & 28.94 & 0.51 & 2.27 \\
    Trellis
    & 29.43 & 21.61 & \second{0.11} & 3.42 \\

    \midrule
    \rowcolor{gray!14}
    \multicolumn{5}{@{}l}{\textbf{\textit{Unified 3D models}}} \\
    Shap.-Omni
    & 27.98 & 24.40 & 0.15 & 3.21 \\
    CoRe3D
    & \second{37.66}
    & \second{20.55}
    & \second{0.11}
    & \second{3.68} \\
    \rowcolor{gray!8}
    \textbf{\modelnamegradient}
    & \best{39.01}
    & \best{16.50}
    & \best{0.10}
    & \best{3.82} \\
    \bottomrule
  \end{tabular}%
  }

  \vspace{-0.55cm}
\end{wraptable}
\noindent\textbf{Text-to-3D Generation.}
\Cref{tab:quant_text_to_3d} shows \modelname{} leading all metrics, improving over CoRe3D by +1.35 CLIP, -4.05 FD, -0.01 KD, and +0.14 Q-Align.
The large improvement in CLIP over Trellis (+9.58) validates that anchor tokens succeed at binding linguistic intent to fine geometric structure. Results also show that unified semantic-geometric reasoning is most valuable when the input is underspecified, and the model must infer object structure from language rather than copy visible image evidence.

\begin{table}[t]
  \centering
  \renewcommand{\arraystretch}{1.0}
  \setlength{\tabcolsep}{7.0pt}
  \caption{\textbf{3D Object Captioning.}
  \best{Best} and \second{second-best} results.
  \modelname{} achieves the strongest performance across metrics, indicating more accurate and semantically grounded 3D descriptions.}
  \label{tab:quant_3d_captioning}
  \vspace{-0.25cm}

  \resizebox{0.99\textwidth}{!}{%
  \begin{tabular}{@{}lccccc@{}}
    \toprule
    \cmidrule(lr){2-4}
    \cmidrule(lr){5-6}
    \textbf{Model} & \textbf{BLEU-1} $\uparrow$
    & \textbf{ROUGE-L} $\uparrow$
    & \textbf{METEOR} $\uparrow$
    & \textbf{Sentence-BERT} $\uparrow$
    & \textbf{SimCSE} $\uparrow$ \\
    \midrule

    \rowcolor{gray!14}
    \multicolumn{6}{@{}l}{\textbf{\textit{General vision-language models}}} \\
    LLaVA-13B
    & 4.01 & 8.18 & 13.18 & 46.97 & 48.86 \\
    Qwen2.5-VL-7B
    & 4.05 & 7.85 & 14.23 & 48.90 & 50.86 \\

    \midrule
    \rowcolor{gray!14}
    \multicolumn{6}{@{}l}{\textbf{\textit{3D-specialized understanding models}}} \\
    3D-LLM
    & 15.11 & 17.84 & 19.22 & 42.36 & 43.58 \\
    LEO
    & 16.98 & 20.12 & 20.91 & 48.01 & 47.25 \\
    PointLLM-13B
    & 3.18 & 7.54 & 12.24 & 47.89 & 49.01 \\

    \midrule
    \rowcolor{gray!14}
    \multicolumn{6}{@{}l}{\textbf{\textit{Unified 3D understanding and generation models}}} \\
    ShapeLLM-Omni
    & 18.92 & 21.46 & 22.12 & 49.43 & 50.72 \\
    CoRe3D
    & \second{24.02}
    & \second{26.45}
    & \second{24.98}
    & \second{51.17}
    & \second{52.79} \\
    \rowcolor{gray!8}
    \textbf{\modelnamegradient}
    & \best{26.22}
    & \best{28.15}
    & \best{25.78}
    & \best{54.73}
    & \best{54.11} \\
    \bottomrule
  \end{tabular}%
  }
  \vspace{-0.1cm}
\end{table}

\noindent\textbf{3D Object Understanding.}
A key promise of unified 3D foundation models is that generative training also strengthens understanding. 
\Cref{tab:quant_3d_captioning} shows that \modelname{} sets a new state-of-the-art across all five metrics, improving over CoRe3D by +2.20 BLEU-1, +1.70 ROUGE-L, +0.80 METEOR, +3.56 Sentence-BERT, and +1.32 SimCSE.
The larger gains on semantic metrics suggest that routing language cues to scale-specific geometric evidence helps the model produce captions that are more semantically aligned with the underlying 3D object.

\begin{figure}[t!]
    \centering
    \includegraphics[width=0.99\linewidth]{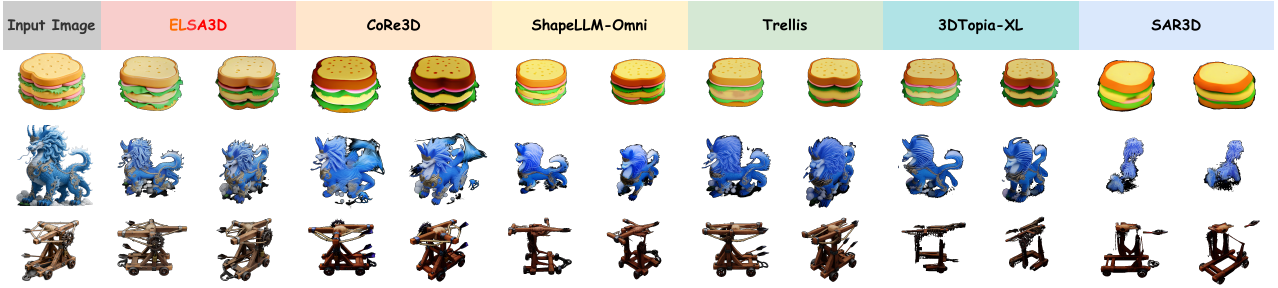}
    \includegraphics[width=0.99\linewidth]{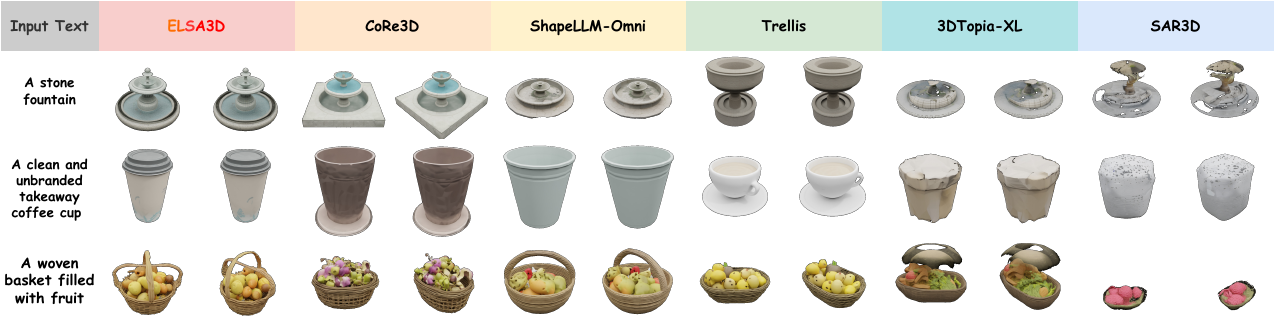}
    \vspace{-0.3cm}
\caption{\textbf{Qualitative Comparison across Image-to-3D and Text-to-3D Generation.} For image-to-3D generation, \modelnamegradient{} better preserves global shape and local appearance cues from the input image. For text-to-3D generation, \modelnamenc{} more faithfully follows category-level intent and fine-grained prompt constraints, including object parts, support structures, materials, and surface appearance.}
    \label{fig:qual_merged}
    \vspace{-0.3cm}
\end{figure}

\noindent\textbf{Qualitative Examples.}
\Cref{fig:qual_merged} compares image-to-3D and text-to-3D generation outputs on visually complex inputs spanning toys, food, vehicles, and stylized characters, showing that \modelname
\begin{wrapfigure}{r}{0.56\textwidth}
    \vspace{-0.2cm}
    \centering
    \includegraphics[width=0.99\linewidth]{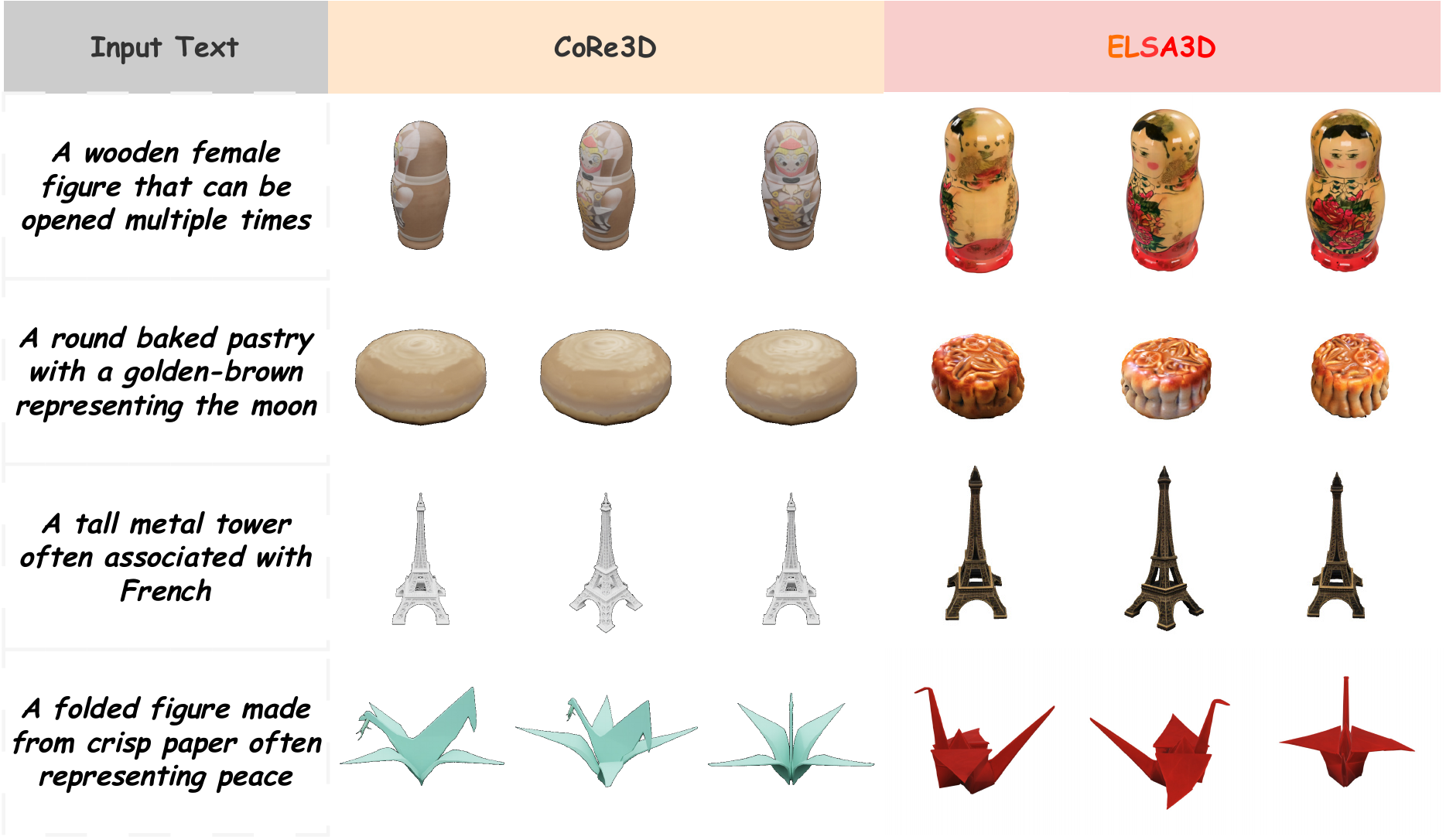}
    \vspace{-0.8cm}
    \caption{\textbf{Reasoning-Based 3D generation.}
    \modelnamegradient{} resolves indirectly described target objects.}
    \label{fig:qual_reasoning_gen}
    \vspace{-0.6cm}
\end{wrapfigure}
consistently produces shapes that retain both the global shape and local appearance cues of the input images, including thin structures, part layout, and distinctive textures.
For text-conditioned 
generation, \modelnamenc{} better satisfies category-level intent and fine-grained prompt constraints, such as object parts, support structures (\eg basket holds fruit), material cues, and surface appearance  (\eg the coffee cup is clean and unbranded as specified).\looseness-1

\noindent\textbf{Reasoning-Based 3D Generation.}
We further evaluate challenging indirect descriptive prompts that do not explicitly name the target object but instead provide {a} description through indirect cues. \Cref{fig:qual_reasoning_gen} shows that \modelname{} infers the intended object identity and generates the corresponding structure and appearance, including a nestable wooden female figure, a moon-associated baked pastry, a French metal tower, and a folded paper symbol of peace with recognizable silhouettes. In contrast, CoRe3D often captures only a coarse or generic shape of the referenced concept. These examples illustrate the value of semantic anchoring under language ambiguity, where the model must recover the latent concept from indirect cues and bind it to geometric evidence at the appropriate scale.

\subsection{Ablations}

\begin{table*}[t!]
  \centering
  \renewcommand{\arraystretch}{1.10}
  \setlength{\tabcolsep}{6pt}
  \caption{\textbf{Ablation on Semantic-Trace Decomposition.}
  We evaluate how each semantic aspect, \emph{Global} (G), \emph{Structure} (S), and \emph{Appearance} (A), contributes to generation, captioning, and general reasoning.
  \emph{No Trace} feeds the raw prompt directly, while \emph{Monolithic} replaces the three-aspect trace with a single free-form reasoning paragraph.
  \best{Best} and \second{second-best} results are highlighted.}
  \label{tab:ablation_semantic_trace}
  \resizebox{\textwidth}{!}{%
  \begin{tabular}{@{}lccc ccc cc cc c@{}}
    \toprule
    \multirow{2}{*}{\textbf{Variant}}
    & \multicolumn{3}{c}{\textbf{Aspects}}
    & \multicolumn{3}{c}{\textbf{Text-to-3D}}
    & \multicolumn{2}{c}{\textbf{Captioning}}
    & \multicolumn{2}{c}{\textbf{General Reasoning}}
    & \multirow{2}{*}{\textbf{Lat. (s)} $\downarrow$} \\
    \cmidrule(lr){2-4}
    \cmidrule(lr){5-7}
    \cmidrule(lr){8-9}
    \cmidrule(lr){10-11}
    & \textbf{G}
    & \textbf{S}
    & \textbf{A}
    & \textbf{CLIP} $\uparrow$
    & \textbf{FD} $\downarrow$
    & \textbf{KD} $\downarrow$
    & \textbf{MET.} $\uparrow$
    & \textbf{SimCSE} $\uparrow$
    & \textbf{MMLU} $\uparrow$
    & \textbf{GSM8K} $\uparrow$
    & \\
    \midrule

    \rowcolor{gray!14}
    \multicolumn{12}{@{}l}{\textbf{\textit{Unstructured prompting baselines}}} \\
    No Trace
    & \xmark & \xmark & \xmark
    & 37.18 & 20.92 & 0.16
    & 23.91 & 51.82
    & 67.3 & 56.1
    & \best{16.1} \\

    Monolithic
    & -- & -- & --
    & 38.32 & 18.21 & 0.13
    & 25.04 & 53.47
    & 67.8 & 57.2
    & 17.2 \\

    \midrule
    \rowcolor{gray!14}
    \multicolumn{12}{@{}l}{\textbf{\textit{Single-aspect traces}}} \\
    G only
    & \cmark & \xmark & \xmark
    & 38.11 & 19.08 & 0.14
    & 24.44 & 52.92
    & 67.7 & 56.9
    & \second{17.1} \\

    S only
    & \xmark & \cmark & \xmark
    & 37.74 & 18.62 & 0.13
    & 24.28 & 52.70
    & 67.6 & 57.0
    & 17.2 \\

    A only
    & \xmark & \xmark & \cmark
    & 37.86 & 19.84 & 0.15
    & 24.52 & 53.04
    & 67.4 & 56.6
    & \second{17.1} \\

    \midrule
    \rowcolor{gray!14}
    \multicolumn{12}{@{}l}{\textbf{\textit{Two-aspect traces}}} \\
    G + S
    & \cmark & \cmark & \xmark
    & 38.64 & 17.31 & 0.12
    & 25.12 & 53.53
    & \second{67.9} & \second{57.6}
    & 17.3 \\

    G + A
    & \cmark & \xmark & \cmark
    & \second{38.72} & 17.89 & 0.12
    & \second{25.31} & \second{53.82}
    & 67.8 & 57.4
    & 17.2 \\

    S + A
    & \xmark & \cmark & \cmark
    & 38.49 & \second{17.08} & \second{0.11}
    & 25.22 & 53.76
    & 67.7 & 57.3
    & 17.2 \\

    \midrule
    \rowcolor{gray!8}
    \textbf{\modelnamegradient}
    & \cmark & \cmark & \cmark
    & \best{39.01} & \best{16.50} & \best{0.10}
    & \best{25.78} & \best{54.11}
    & \best{68.1} & \best{58.2}
    & 17.2 \\
    \bottomrule
  \end{tabular}%
  }

\end{table*}
\noindent\textbf{Semantic Trace Decomposition.}
We test the contribution of decomposing the semantic trace into Global (G), Structure (S), and Appearance (A) aspects (Table~\ref{tab:ablation_semantic_trace}). Removing the trace entirely (\textit{No Trace}) is the weakest configuration (Text-to-3D CLIP 37.18). Replacing the structured decomposition with a single free-form reasoning paragraph of comparable length (\textit{Monolithic}) recovers part of the gain (CLIP 38.32) but still trails the structured variant, indicating that the gain comes from the decomposition itself rather than from added tokens. Among subset combinations, the full G+S+A trace is best across all quality metrics. The pattern matches the scale-routing inductive bias: global semantics, structural semantics, and appearance semantics naturally bind to corresponding geometric scales respectively, and removing any aspect deprives anchors at the corresponding scale of clean input.\looseness-1

\begin{table*}[t!]
  \vspace{-4pt}
  \centering
  \renewcommand{\arraystretch}{1.10}
  \setlength{\tabcolsep}{4pt}
  \caption{\textbf{Ablation on Elastic Computation.}
  We evaluate dynamic block skipping and adaptive MLP width. 
  \emph{Full-Compute} executes every block at full width and is reported as an unaccelerated upper-bound reference, excluded from best/second-best highlighting. 
  \best{Best} and \second{second-best} results.}
  \label{tab:ablation_elastic_designs}
  \resizebox{\textwidth}{!}{%
  \begin{tabular}{@{}lccc ccc cc cc@{}}
    \toprule
    \multirow{2}{*}{\textbf{Variant}}
    & \multicolumn{3}{c}{\textbf{Text-to-3D}}
    & \multicolumn{3}{c}{\textbf{Image-to-3D}}
    & \multicolumn{2}{c}{\textbf{Captioning}}
    & \multicolumn{2}{c}{\textbf{Compute}} \\
    \cmidrule(lr){2-4}
    \cmidrule(lr){5-7}
    \cmidrule(lr){8-9}
    \cmidrule(lr){10-11}
    & \textbf{CLIP} $\uparrow$
    & \textbf{FD} $\downarrow$
    & \textbf{KD} $\downarrow$
    & \textbf{CLIP} $\uparrow$
    & \textbf{FD} $\downarrow$
    & \textbf{KD} $\downarrow$
    & \textbf{MET.} $\uparrow$
    & \textbf{SimCSE} $\uparrow$
    & \textbf{FLOPs (G)} $\downarrow$
    & \textbf{Lat. (s)} $\downarrow$ \\
    \midrule

    \rowcolor{gray!8}
    Full-Compute \textit{(reference)}
    & 39.18 & 16.22 & 0.09
    & 89.34 & 8.92 & 0.06
    & 25.89 & 54.28
    & 1284 & 34.6 \\

    \midrule
    Depth-Only Elastic
    & \second{38.52} & \second{17.61} & \second{0.12}
    & \second{88.46} & \second{10.42} & \second{0.08}
    & \second{25.23} & \second{53.64}
    & \second{811} & \second{24.1} \\

    Width-Only Elastic
    & 38.21 & 18.04 & \second{0.12}
    & 88.12 & 10.91 & \second{0.08}
    & 25.05 & 53.37
    & 985 & 27.0 \\

    \rowcolor{gray!8}
    \textbf{\modelnamegradient}
    & \best{39.01} & \best{16.50} & \best{0.10}
    & \best{89.21} & \best{9.23} & \best{0.06}
    & \best{25.78} & \best{54.11}
    & \best{632} & \best{17.2} \\
    \bottomrule
  \end{tabular}%
  }

\end{table*}
\noindent\textbf{Elastic Computation Designs.}
We evaluate whether dynamic block skipping and adaptive MLP width preserve quality while reducing compute (Table~\ref{tab:ablation_elastic_designs}). \textit{Full-Compute} runs every block at full width and serves as an upper bound (Text-to-3D CLIP 39.18, {Text-to-3D FD 16.22}, but at 1284G FLOPs and 34.6s latency). Depth-only and width-only elastic variants reduce cost partially, but each loses noticeable quality (Text-to-3D CLIP drops to 38.52 and 38.21, respectively). \modelname combines both axes and recovers near-upper-bound quality (CLIP 39.01, FD 16.50) while halving FLOPs (632G) and cutting latency by more than 2$\times$ (17.2s). This confirms the effectiveness of our elastic compute design.\looseness-1

\begin{table*}[t]
  \centering
  \renewcommand{\arraystretch}{1.10}
  \setlength{\tabcolsep}{4.2pt}
  \caption{\textbf{Anchor Token Ablation.}
  \modelname{} achieves the best performance across all metrics while remaining substantially more efficient than dense cross-modal fusion.
  \best{Best} and \second{second-best} results.}
  \label{tab:ablation_anchors}
  \resizebox{0.99\textwidth}{!}{%
  \begin{tabular}{@{}lccc ccc cc cc@{}}
    \toprule
    \multirow{2}{*}{\textbf{Variant}}
    & \multicolumn{3}{c}{\textbf{Text-to-3D}}
    & \multicolumn{3}{c}{\textbf{Image-to-3D}}
    & \multicolumn{2}{c}{\textbf{Captioning}}
    & \multicolumn{2}{c}{\textbf{Cost}} \\
    \cmidrule(lr){2-4}
    \cmidrule(lr){5-7}
    \cmidrule(lr){8-9}
    \cmidrule(lr){10-11}
    & \textbf{CLIP} $\uparrow$
    & \textbf{FD} $\downarrow$
    & \textbf{KD} $\downarrow$
    & \textbf{CLIP} $\uparrow$
    & \textbf{FD} $\downarrow$
    & \textbf{KD} $\downarrow$
    & \textbf{MET.} $\uparrow$
    & \textbf{SimCSE} $\uparrow$
    & \textbf{FLOPs (G)} $\downarrow$
    & \textbf{Lat. (s)} $\downarrow$ \\
    \midrule

    No Anchors
    & 36.72 & 21.84 & 0.17
    & 85.94 & 14.78 & 0.12
    & 23.42 & 50.87
    & \best{568} & \best{15.4} \\

    Direct Cross-Attn
    & 38.32 & \second{17.91} & 0.12
    & 88.41 & \second{10.72} & 0.08
    & 24.96 & 53.34
    & 1081 & 29.8 \\

    Dense Anchors
    & \second{38.57} & 18.34 & \second{0.11}
    & \second{88.63} & 11.16 & \second{0.07}
    & \second{25.21} & \second{53.62}
    & 865 & 23.6 \\

    \rowcolor{gray!8}
    \textbf{\modelnamegradient}
    & \best{39.01} & \best{16.50} & \best{0.10}
    & \best{89.21} & \best{9.23} & \best{0.06}
    & \best{25.78} & \best{54.11}
    & \second{632} & \second{17.2} \\
    \bottomrule
  \end{tabular}%
  }

  \vspace{-0.3cm}
\end{table*}
\noindent\textbf{Anchor Tokens.}
We next ablate the mechanism used to connect semantic and geometric tokens.
\Cref{tab:ablation_anchors} compares four designs: removing anchors entirely (\textit{No Anchors}), replacing anchors with full bidirectional cross-attention between all text and 3D tokens (\textit{Direct Cross-Attn}), disabling routing and instantiating dense anchors for every text token (\textit{Dense Anchors}), and the proposed elastic anchors.
Removing anchors degrades every task, confirming that implicit self-attention over a flat sequence is insufficient for reliable text--3D alignment.
Direct cross-attention recovers quality but requires 1081G FLOPs and 29.8s latency, while dense anchors still cost 865G FLOPs and 23.6s.
\modelname{} obtains the best quality across generation and captioning while using only 632G FLOPs and 17.2s latency, showing that sparse anchor selection is not just cheaper but also less noisy than dense fusion.

\begin{table*}[t]
  \centering
  \renewcommand{\arraystretch}{1.10}
  \setlength{\tabcolsep}{8pt}
  \caption{\textbf{Scale-Aware Anchor Routing Ablation.}
  Learned routing achieves the best quality across text-to-3D generation, image-to-3D generation, and 3D captioning while remaining close to the fastest single-scale variant.
  \best{Best} and \second{second-best} results are highlighted.}
  \label{tab:ablation_scale_routing}
  \resizebox{0.99\textwidth}{!}{%
  \begin{tabular}{@{}lccc ccc cc c@{}}
    \toprule
    \multirow{2}{*}{\textbf{Variant}}
    & \multicolumn{3}{c}{\textbf{Text-to-3D}}
    & \multicolumn{3}{c}{\textbf{Image-to-3D}}
    & \multicolumn{2}{c}{\textbf{Captioning}}
    & \multirow{2}{*}{\textbf{Lat. (s)} $\downarrow$} \\
    \cmidrule(lr){2-4}
    \cmidrule(lr){5-7}
    \cmidrule(lr){8-9}
    & \textbf{CLIP} $\uparrow$
    & \textbf{FD} $\downarrow$
    & \textbf{KD} $\downarrow$
    & \textbf{CLIP} $\uparrow$
    & \textbf{FD} $\downarrow$
    & \textbf{KD} $\downarrow$
    & \textbf{MET.} $\uparrow$
    & \textbf{SimCSE} $\uparrow$
    & \\
    \midrule

    All-Scale Attn.
    & \second{38.72} & \second{17.21} & \second{0.11}
    & \second{88.71} & \second{10.18} & \second{0.07}
    & \second{25.36} & \second{53.76}
    & 22.4 \\

    Coarse-Only
    & 37.84 & 19.76 & 0.14
    & 86.92 & 13.40 & 0.10
    & 24.46 & 52.41
    & \best{16.5} \\

    Fine-Only
    & 38.09 & 18.88 & 0.13
    & 87.63 & 12.46 & 0.09
    & 24.73 & 52.86
    & 18.6 \\

    Random Scale
    & 35.31 & 21.24 & 0.19
    & 81.05 & 14.82 & 0.14
    & 22.02 & 49.21
    & 17.6 \\

    \rowcolor{gray!8}
    \textbf{\modelnamegradient}
    & \best{39.01} & \best{16.50} & \best{0.10}
    & \best{89.21} & \best{9.23} & \best{0.06}
    & \best{25.78} & \best{54.11}
    & \second{17.2} \\
    \bottomrule
  \end{tabular}%
  }

  \vspace{-0.3cm}
\end{table*}
\noindent\textbf{Scale-Aware Anchor Routing.}
We next test whether anchors must be routed to a learned geometric scale.
\Cref{tab:ablation_scale_routing} shows that random scale assignment is the weakest variant, indicating that scale diversity alone is not enough.
Coarse-only and fine-only routing also underperform because they force all semantic cues into a single resolution: category-level information benefits from coarse geometry, while part and appearance cues often require finer scales.
All-scale attention is closer in quality but increases latency from 17.2s to 22.4s because every anchor attends to all octree scales.
The learned router achieves the best quality with near coarse-only latency, validating scale-aware anchoring as a useful inductive bias for coarse-to-fine 3D grounding.

\section{Conclusion}
We propose \modelname{}, a unified 3D understanding-and-generation model that makes language--geometry interaction sparse, adaptive, and scale-aware. \modelname{} introduces semantic anchor tokens to route selected language cues to the most relevant level of a multiscale 3D hierarchy and write the fused evidence back into the shared representation. A lightweight router jointly controls semantic grounding, block execution, and MLP width, allocating cross-modal capacity and computation according to the input. Across image-to-3D, text-to-3D, reasoning-based generation, and 3D captioning, \modelname{} consistently outperforms strong 3D and unified baselines. Ablations demonstrate the benefits of semantic decomposition, sparse anchors, and learned scale selection, while elastic computation retains near-full-compute quality at roughly half the FLOPs and inference latency. These findings highlight the value of coupling explicit language--geometry grounding with adaptive computation in unified 3D models. Future work will explore extensions to multi-object scenes and interactive editing.\looseness-1

\begin{ack}
This research was partially supported by the NSF CAREER Award \#2542328 and the Lambda Research Grant. The views and conclusions contained herein are those of the authors and should not be interpreted as necessarily representing the official policies, either expressed or implied, of NSF, Lambda, or the U.S. Government. The U.S. Government is authorized to reproduce and distribute reprints for governmental purposes notwithstanding any copyright annotation therein. 
\end{ack}


\bibliographystyle{plain}
\bibliography{neurips_2026}

\newpage
\appendix

\section{Related Work}\label{app:related}

\textbf{Hierarchical 3D Tokenization and Generation.}
{Recent 3D generative models increasingly replace flat dense grids with compact structured representations that expose sparsity, hierarchy, or progressive detail~\cite{zhang2024clay,xiong2025octfusion,dutt2026lost,zhang2025vertexregen,yu2026silsa}.} Orthogonal to native 3D tokenizers, optimization-based text- and image-to-3D methods distill pretrained 2D diffusion priors into 3D representations~\cite{poole2022dreamfusion,qian2023magic123,lin2023magic3d,wang2023prolificdreamer,chen2023fantasia3d,raj2023dreambooth3d,sweetdreamer,sun2023dreamcraft3d,chen2024text,sjc,tang2023dreamgaussian,yi2024gaussiandreamer}.
These approaches provide strong specialized baselines for text- or image-conditioned 3D generation, but typically require per-instance optimization and do not form a single autoregressive backbone for both 3D understanding and generation. {Complementary approaches explore feed-forward reconstruction, multiview diffusion, and latent- or mesh-based 3D generation~\cite{hong2023lrm,liu2023zero,liu2023syncdreamer,long2024wonder3d,tang2024lgm,siddiqui_meshgpt_2024,zhao2023michelangelo,wang2023rodin,wu2024direct3d,yang2024hunyuan3d,huang2025spar3d,chen20253dtopia,li2024craftsman,ye2025hi3dgen}.}
Octree-based models exploit spatial subdivision to scale 3D perception, reconstruction, and generation~\cite{wang2023octformer,wei2025octgpt,xiong2025octfusion,yu2021plenoctrees,szeliski1993rapid}, with extensions to textured assets~\cite{liu2024texoct,xiong2025texgaussian,benson2002octree}.
{Other approaches pursue flow-based 3D shape tokenization~\cite{chang20243d}, multi-resolution latent diffusion~\cite{zhang2024clay}, mesh autoregressive detailization~\cite{gao2025mars}, semantic token ordering~\cite{dutt2026lost}, and continuous levels of detail~\cite{zhang2025vertexregen}. Memory-bounded training, representation simplification, and part-aware modeling are also studied for 3D Gaussian splatting~\cite{zhang2026gaussians,zhang2025gaussianspa,yu2026part2gs}. These works organize or compress geometric representations. \modelname{} additionally uses the hierarchy as a cross-modal interface, routing language-derived semantic anchors to explicit geometric scales.}

\textbf{3D-Language Models and Unified 3D Understanding-Generation.}
{Recent 3D multimodal models connect language with 3D representations for question answering, captioning, grounding, dialogue, and scene understanding~\cite{hong20233d,xu2024pointllm,qi2024shapellm,wang2023chat,chen2024ll3da,xue2023ulip,fu2024scene,wang2024open}. More recent systems extend this direction toward joint 3D understanding and generation~\cite{ye2025shapellm_omni,xu2026uniugg,yu2025core3d,chen2025sar3d,wang2024llama,liu2024uni3d}, while DreamPartGen studies semantically grounded part-level 3D generation~\cite{yu2026dreampartgen}. These motivate language-driven 3D modeling. Our focus is the explicit routing of semantic tokens to geometric resolutions.}
\modelname{} exposes the hierarchical structure of 3D geometry to the language model and grounds selected semantic tokens through scale-specific anchor units, enabling global, part-level, and local evidence to be fused within a unified autoregressive representation.

\textbf{Adaptive Cross-Modal Routing.}
{Efficiency methods combine visual-token reduction~\cite{bolyatoken,liang2022not,zhang2025vqtoken,chang2026d3tom}, token selection, multimodal fusion, and cache compression~\cite{wang2022multimodal,zhang2024magic,zhong2025aim,li2025flowmm,wang2026fasa}, query-based connectors~\cite{li2023blip,alayrac2022flamingo}, and cross-modal token alignment~\cite{xu2022groupvit,yin2025sea,zhang2025attanchor,susladkar2026pyratok}. Adaptive computation further adjusts block execution and token budgets~\cite{zhao2024dynamic,wang2026elastic,zhu2025ea,wang2026pare,cheng2026training,shen2026latteflow}, expert allocation~\cite{shen2026cogniroute}, or trajectory-context length~\cite{wu2023elastic}. Complementary approaches use low-rank compression and quantization~\cite{wang2026qsvd,wang2026wsvd,wang2026laser,hong2026safe,zhang2026not}, efficient adaptation~\cite{xiao2026not}, and video-generation distillation~\cite{wang2026vdot,wang2026vdot++}. \modelname{} couples adaptive computation with grounding by routing selected semantic anchors to scale-specific 3D features and returning the fused evidence to the unified representation.}

\textbf{Spatial Reasoning, World Models, and Controlled Generation and Learning.}
Broader connections include spatial localization and reasoning~\cite{yu2025point2rbox,zhang2026point2rbox,zhang2025pointobb,lu2026mllms,shen2025spatialreasoner}, world modeling~\cite{zhang2026thinkjepa,dong2026language,dong2026towards,shen2026phantom}, and embodied perception, reasoning, and control~\cite{dong2025ha,liu2026palm,wang2026towermind,wang2026mars,liu2026driveva,liu2026universe,liu2026streamvlo}, with benchmarks and surveys covering 3D world synthesis and physical AI~\cite{lu2026worldcoder,zhang2026survey}. Cross-modal learning finds diverse applications~\cite{chi2022missing,feng2026triemo,hu2026meow,feng2025unveiling,lan2026braintrace,zhang2026dive,zhang2026linkedout,chi2026enhancing_video,liu2026reasonact,nguyen2025calico}, alongside work on adaptation and in-context learning~\cite{li2026personalize,li2025m,li2026make,li2026acticl,xiao2026adapting} and reliability~\cite{zhang2026evaluating,xia2025rethinking,zhang2026devils,ogunleye2026vcd}. Related efforts also explore controlled generation across modalities~\cite{tian2024var,zhou2025panollama,lu2025uni,lu2026one,huang2026dramaagent,zhang2026active,susladkar2026best,venkatesh2026ravel,shah2026graphvid}, language-model training~\cite{zhang2026consolidating,ye2026reconsidering,ye2026sumi}, graph representations~\cite{zeng2026dta,zhang2026heteromile}, and compressed latent reasoning~\cite{chang2026struggle}. Inspired by these broader advances in structured representation and cross-modal learning, \modelname{} grounds semantic concepts in scale-specific geometric evidence for unified 3D understanding and generation.

\section{Training and Inference Details}\label{app:training}
\textbf{Stage 1: Octree-Based 3D VQ-VAE Training.}
The octree-based VQ-VAE consists of an encoder, scale-specific codebooks, and a decoder. It is trained on 3D data alone, without text supervision. The encoder maps each $128^3$ voxel grid to the multiscale octree representation described in \Cref{sec:representations}, whereas the decoder reconstructs the voxel occupancy from the predicted octree topology and quantized content embeddings. Training minimizes a reconstruction loss together with the standard VQ commitment and codebook losses~\cite{van2017neural}:
\begin{equation}
\setlength{\abovedisplayskip}{3pt}
\setlength{\belowdisplayskip}{3pt}
\setlength{\abovedisplayshortskip}{2pt}
\setlength{\belowdisplayshortskip}{2pt}
    \mathcal{L}_{\mathrm{VQ}} 
    = \mathcal{L}_{\mathrm{recon}} 
    + \lambda_{\mathrm{commit}}
      \sum_{v}\|\mathbf{f}_v - \mathrm{sg}(\mathbf{c}_v)\|_2^2
    + \lambda_{\mathrm{code}}
      \sum_{v}\|\mathrm{sg}(\mathbf{f}_v) - \mathbf{c}_v\|_2^2,
\end{equation}
where $\mathbf{f}_v$ is the encoder feature at node $v$, $\mathbf{c}_v$ is its assigned codebook embedding, and $\mathrm{sg}(\cdot)$ denotes stop-gradient. After this stage, the VQ-VAE encoder and codebooks are frozen.

At inference time for 3D generation, the structural bits generated by the unified model are consumed deterministically to rebuild the octree topology. For each existing node, the predicted content code $k_v$ is mapped to its codebook embedding $\mathbf{g}_v=\mathcal{C}^{(s)}[k_v]$ and placed at the corresponding octree location. The VQ-VAE decoder then reconstructs a dense $128^3$ voxel grid, from which we obtain a textured mesh using a texture transformer~\cite{xiang2025structured} and mesh decoder.

\textbf{Stage 2: Unified Autoregressive Training.}
We train the unified model with autoregressive next-token prediction over a combined vocabulary spanning text tokens, structural-bit tokens, and 3D tokens. Following ShapeLLM-Omni~\cite{ye2025shapellm_omni}, we extend the base model's vocabulary with the VQ-VAE codebook entries and two structural-bit tokens. Let $\mathbf{z}=(z_1,\dots,z_T)$ denote the resulting training sequence. The autoregressive objective is
\begin{equation}
\setlength{\abovedisplayskip}{3pt}
\setlength{\belowdisplayskip}{3pt}
\setlength{\abovedisplayshortskip}{2pt}
\setlength{\belowdisplayshortskip}{2pt}
    \mathcal{L}_{\mathrm{AR}} 
    = -\sum_{t=1}^{T} \log p_\theta(z_t \mid z_{<t}),
\end{equation}
where $z_t$ may be a text token, a structural bit $o_v\in\{0,1\}$, or a discrete 3D content code $k_v$. The corresponding codebook embedding $\mathbf{g}_v=\mathcal{C}^{(s)}[k_v]$ is then augmented with positional and scale signals before entering the transformer.

\textbf{Depth and Width Budgets.}
Without explicit pressure, the router learns to execute every block at full width. We introduce two budget losses. Let $\bar{p}=\tfrac{1}{n}\sum_{i=1}^{n} p^{i}$ denote the 
mean execution probability and let $ r^i\!=\!\sum_{j=1}^{4} q_j^i \mathcal{W}_j$ denote 
the expected width of block $i$, averaged only over non-skipped blocks:
\begin{equation}
    \bar{r} = \frac{\sum_{i=1}^{n}
    \mathbf{1}[p^{i}\ge\tau]\, r^{i}}
    {\sum_{i=1}^{n}\mathbf{1}[p^{i}\ge\tau]}.
\end{equation}
Given target compute budgets $\rho_d,\rho_w\in(0,1)$, we penalize deviations between the router's realized average depth and width usage and their desired budgets:
\begin{equation}
    \mathcal{L}_{\text{depth}} = (\bar{p} - \rho_d)^2,
    \qquad
    \mathcal{L}_{\text{width}} = (\bar{r} - \rho_w)^2.
\end{equation}
These terms prevent the router from collapsing to full computation while allowing the realized compute to remain input-dependent.

\textbf{Anchor Sparsity and Scale Diversity.}
To keep anchor construction compact, we $\ell_1$-regularize the anchor gates, whereas to prevent scale routing from collapsing to a single resolution, we minimize the negative entropy of the scale distribution:
\begin{equation}
    \mathcal{L}_{\mathrm{sparse}} 
    = \frac{1}{nM}\sum_{i=1}^{n}\sum_{m=1}^{M} 
    |\beta_m^i|,
    \qquad
    \mathcal{L}_{\mathrm{scale}} 
    = \frac{1}{nM}\sum_{i=1}^{n}\sum_{m=1}^{M}
    \sum_{s=1}^{S} \pi_{m,s}^{i}\log \pi_{m,s}^{i}.
\end{equation}

All auxiliary losses are computed from quantities already produced by the router and add negligible overhead. 
Gradient 
flow through the three discrete router decisions (block 
gate, width selection, scale choice) is handled by 
straight-through estimators~\cite{bengio2013ste} in all 
cases. 
Crucially, the write-back mechanism (\Cref{sec:alignment}) allows $\mathcal{L}_{\text{AR}}$ to back-propagate through anchor construction
 into the router's anchor-selection and scale-selection heads, so these heads receive task-level supervision in addition to the auxiliary budget losses.
The final training objective is
\begin{equation}
\setlength{\abovedisplayskip}{3pt}
\setlength{\belowdisplayskip}{2pt}
\setlength{\abovedisplayshortskip}{2pt}
\setlength{\belowdisplayshortskip}{2pt}
    \mathcal{L}
    =
    \mathcal{L}_{\mathrm{AR}}
    + \lambda_d \mathcal{L}_{\mathrm{depth}}
    + \lambda_w \mathcal{L}_{\mathrm{width}}
    + \lambda_s \mathcal{L}_{\mathrm{sparse}}
    + \lambda_c \mathcal{L}_{\mathrm{scale}}.
\end{equation}

\textbf{Inference.}
At inference time, blocks with $p^i<\tau$ are skipped entirely, and no anchors or write-back are computed for those blocks. For executed blocks, the selected MLP width is materialized by weight slicing:
\begin{equation}
    \mathrm{MLP}_{\hat{w}^{i}}(\mathbf{z})
    =
    \sigma\!\left(
    \mathbf{z}\mathbf{W}_{1}[:,:\hat{w}^{i}H]
    \right)
    \mathbf{W}_{2}[:\hat{w}^{i}H,:],
\end{equation}
and anchors are instantiated only for semantic tokens with $\beta_m^i\ge\tau_a$, drawing geometric evidence from scale
\( 
    \alpha_m^i\!=\!\arg\max_{s}\pi_{m,s}^i.
\)
Because block execution, width, anchor selection, and scale assignment are all input-dependent, the realized computation and cross-modal interaction adapt to each sample.

\section{Implementation Details}
\label{app:implement_details}
We implement the 3D tokenizer as an OctGPT-style octree VQ-VAE~\cite{wei2025octgpt} with maximum depth $S=7$, 8192-entry scale-specific codebooks, and embedding dimension $C=256$. It is trained for 100 epochs with AdamW, learning rate $1\times10^{-4}$, batch size 32, on 8 NVIDIA A100 GPUs, after which the encoder and codebooks are frozen. The unified model is initialized from {Qwen2.5-VL-7B-Instruct} ($n=28$, $D=3584$)~\cite{bai2025qwen25vl}, following ShapeLLM-Omni~\cite{ye2025shapellm_omni}. We extend the vocabulary with $8192\times S$ 3D tokens and two structural-bit tokens, use a scale tag of dimension 32, and keep the visual encoder frozen. The router uses $d_r=128$, the anchor MLP is a two-layer GELU network with hidden size 512, and anchor write-back uses 4-head cross-attention. We set $\rho_d=0.7$, $\rho_w=0.75$, and $\lambda_d=1.0$, $\lambda_w=1.0$, $\lambda_s=0.01$, $\lambda_c=0.01$. For training data, we follow the 3D-Alpaca construction pipeline of ShapeLLM-Omni~\cite{ye2025shapellm_omni}, which covers text-to-3D, image-to-3D, 3D captioning, and 3D editing tasks. We train on the publicly available 3D-Alpaca dataset~\cite{ye2025shapellm_omni} and supplement it with additional 3D assets from {Trellis-500K}~\cite{xiang2025structured}. We also include UltraChat~\cite{ding2023ultrachat} to preserve general language capabilities. The unified model is trained for 200k steps with AdamW, learning rate decayed from $5\times10^{-5}$ to $5\times10^{-6}$, per-GPU batch size 2 with 4-step gradient accumulation, on 8 NVIDIA A100 GPUs. At inference, we use top-$k=8192$, top-$p=0.7$, temperature $0.7$, and set both routing thresholds to 0.5.

\section{Additional Experiments and Ablations}

\noindent\textbf{General Conversational Capabilities.}
Adding 3D generation and understanding should not come at the cost of the model's original reasoning ability. For language and multimodal reasoning, we compare against general-purpose VLMs, including Qwen2.5-VL-7B~\cite{bai2025qwen25vl} and LLaMA3.2-Vision-11B~\cite{touvron2023llama}, and 3D-focused multimodal models, including LLaMA-Mesh-8B~\cite{wang2024llama}, CoRe3D~\cite{yu2025core3d}, and ShapeLLM-Omni-7B~\cite{ye2025shapellm_omni}.
\Cref{tab:quant_general_conv} shows that \modelname{} remains competitive with general-purpose VLMs while improving over 3D-focused baselines on MMLU, GSM8K, and SIQA.
It achieves the best score on MMLU, GSM8K, and SIQA, and the second-best score on PIQA behind Qwen2.5-VL-7B.
These results suggest that UltraChat mixing and elastic routing preserve broad language competence while adding 3D-specific capabilities.
\begin{table*}[t]
  \centering
  \renewcommand{\arraystretch}{1.10}
  \caption{\textbf{General Conversational and Reasoning Ability.}
  \modelname{} preserves broad language and reasoning capabilities.
  \best{Best} and \second{second-best} results are highlighted.}
  \label{tab:quant_general_conv}
  \resizebox{0.99\textwidth}{!}{%
  \begin{tabular}{@{}lcccccc@{}}
    \toprule
    \multirow{2}{*}{\textbf{Benchmark}}
    & \multicolumn{2}{c}{\textbf{General VLMs}}
    & \textbf{Mesh LLM}
    & \multicolumn{3}{c}{\textbf{Unified 3D Models}} \\
    \cmidrule(lr){2-3}
    \cmidrule(lr){4-4}
    \cmidrule(lr){5-7}
    & \textbf{Qwen2.5-VL}
    & \textbf{LLaMA3.2-Vision}
    & \textbf{LLaMA-Mesh}
    & \textbf{ShapeLLM-Omni}
    & \textbf{CoRe3D}
    & \textbf{\modelnamegradient} \\
    \midrule

    MMLU $\uparrow$
    & 67.5
    & 66.2
    & 59.8
    & 64.3
    & \second{67.6}
    & \best{68.1} \\

    PIQA $\uparrow$
    & \best{81.3}
    & 80.1
    & 79.8
    & 78.9
    & 79.4
    & \second{80.6} \\

    GSM8K $\uparrow$
    & 43.2
    & 42.1
    & 37.2
    & 55.6
    & \second{57.3}
    & \best{58.2} \\

    SIQA $\uparrow$
    & 41.0
    & 40.6
    & 40.3
    & \second{41.5}
    & \second{41.5}
    & \best{41.8} \\
    \bottomrule
  \end{tabular}%
  }

  \vspace{-0.2cm}
\end{table*}

\begin{wraptable}{r}{0.55\textwidth}
    \vspace{-0.3cm}
    \centering
    \caption{\textbf{Semantic-to-scale routing.} Mean routing mass (\%); bold marks each row's maximum. Totals may differ from 100\% due to rounding.}
    \label{tab:semantic_scale_distribution}
    \footnotesize
    \setlength{\tabcolsep}{2pt}
    \renewcommand{\arraystretch}{1.10}
    \begin{tabular*}{\linewidth}{@{\extracolsep{\fill}}llccc@{}}
        \toprule
        \textbf{Aspect} & \textbf{Example tokens} & \textbf{Coarse} & \textbf{Mid} & \textbf{Fine} \\
        & & $s\!\in\!\{1,2\}$ & {$s\!\in\!\{3,4,5\}$} & $s\!\in\!\{6,7\}$ \\
        \midrule
        Global & \shortstack[l]{castle, vehicle,\\upright, silhouette} & \textbf{78.8} & 15.4 & 5.9 \\
        Structure & \shortstack[l]{towers, wheels,\\wings, armrest} & 16.4 & \textbf{51.6} & 31.9 \\
        Appearance & \shortstack[l]{metallic, wooden,\\tiled, studded} & 4.1 & 21.2 & \textbf{74.7} \\
        \bottomrule
    \end{tabular*}
    \vspace{-8pt}
\end{wraptable}
\noindent\textbf{Semantic-to-Scale Routing.}
We average $\pi_{m,s}^{i}$ over all executed blocks and anchor-selected token occurrences in the evaluation set, grouping tokens by semantic aspect and summing probabilities within each scale group.
\Cref{tab:semantic_scale_distribution} shows that Global tokens favor coarse scales (78.8\%), Structure tokens intermediate scales (51.6\%), and Appearance tokens fine scales (74.7\%).
This progression matches global shape, part layout, and surface detail, providing direct evidence that the learned router captures semantic-to-geometric correspondence without manually specified token-to-scale assignments.

\noindent\textbf{Number of Octree Scales.} 
In Table~\ref{tab:ablation_num_scales}, we sweep the maximum octree depth $S \in \{4, 5, 6, 7, 8\}$, varying the finest voxel resolution from $16^3$ to $256^3$. Reconstruction quality improves monotonically with depth (IoU 0.672 $\to$ 0.902, CD
0.049 $\to$ 0.011), but downstream generation saturates at $S{=}7$: pushing to $S{=}8$ marginally regresses Text-to-3D quality while inflating latency by 50\% (25.8s vs.\ 17.2s). The gap between reconstruction and generation curves reflects a known tension in autoregressive 3D modeling~\cite{wei2025octgpt}: deeper octrees produce sequences too long for the unified backbone to model coherently, so tokenizer fidelity outruns the AR model's ability to exploit it. We adopt $S{=}7$ as the best quality-cost trade-off, as $S \geq 9$ also becomes prohibitively expensive. 
\begin{table*}[t]
  \centering
  \renewcommand{\arraystretch}{1.10}
  \setlength{\tabcolsep}{6pt}
  \caption{\textbf{Ablation on the number of octree scales.}
  Increasing the maximum octree scale improves reconstruction fidelity, but very fine scales introduce higher latency and slightly weaker downstream generation.
  \best{Best} and \second{second-best} results are highlighted.}
  \label{tab:ablation_num_scales}
  \resizebox{\textwidth}{!}{%
  \begin{tabular}{@{}cc cc ccc ccc c@{}}
    \toprule
    \multirow{2}{*}{\textbf{$S$}}
    & \multirow{2}{*}{\textbf{Resolution}}
    & \multicolumn{2}{c}{\textbf{Reconstruction}}
    & \multicolumn{3}{c}{\textbf{Text-to-3D}}
    & \multicolumn{3}{c}{\textbf{Image-to-3D}}
    & \multirow{2}{*}{\textbf{Lat. (s)} $\downarrow$} \\
    \cmidrule(lr){3-4}
    \cmidrule(lr){5-7}
    \cmidrule(lr){8-10}
    &
    & \textbf{IoU} $\uparrow$
    & \textbf{CD} $\downarrow$
    & \textbf{CLIP} $\uparrow$
    & \textbf{FD} $\downarrow$
    & \textbf{KD} $\downarrow$
    & \textbf{CLIP} $\uparrow$
    & \textbf{FD} $\downarrow$
    & \textbf{KD} $\downarrow$
    & \\
    \midrule

    4 & $16^3$
    & 0.672 & 0.049
    & 34.68 & 27.42 & 0.25
    & 82.31 & 18.74 & 0.18
    & \best{9.4} \\

    5 & $32^3$
    & 0.792 & 0.024
    & 36.81 & 21.64 & 0.17
    & 85.64 & 14.38 & 0.12
    & \second{11.6} \\

    6 & $64^3$
    & 0.836 & 0.018
    & 38.14 & 18.72 & 0.13
    & 87.92 & 11.67 & 0.09
    & 14.0 \\

    \rowcolor{gray!8}
    \textbf{7} & $\mathbf{128^3}$ \textbf{(\modelnamegradient)}
    & \second{0.864} & \second{0.013}
    & \best{39.01} & \best{16.50} & \best{0.10}
    & \best{89.21} & \second{9.23} & \best{0.06}
    & 17.2 \\

    8 & $256^3$
    & \best{0.902} & \best{0.011}
    & \second{38.74} & \second{17.28} & \second{0.11}
    & \second{89.17} & \best{9.01} & \best{0.06}
    & 25.8 \\

    \bottomrule
  \end{tabular}%
  }

  \vspace{-0.3cm}
\end{table*}
\begin{table*}[t]
  \centering
  \renewcommand{\arraystretch}{1.10}
  \setlength{\tabcolsep}{10pt}
  \caption{\textbf{Ablation on Scale-Specific Codebooks.}
  We compare scale-specific VQ codebooks with a single shared codebook across all octree scales.
  Scale-specific codebooks improve both VQ-VAE reconstruction and downstream 3D generation, suggesting that each scale benefits from a specialized geometric vocabulary.
  \best{Best} results are highlighted.}
  \label{tab:ablation_codebook}
  \resizebox{0.99\textwidth}{!}{%
  \begin{tabular}{@{}lcc ccc ccc@{}}
    \toprule
    \multirow{2}{*}{\textbf{Variant}}
    & \multicolumn{2}{c}{\textbf{Reconstruction}}
    & \multicolumn{3}{c}{\textbf{Text-to-3D}}
    & \multicolumn{3}{c}{\textbf{Image-to-3D}} \\
    \cmidrule(lr){2-3}
    \cmidrule(lr){4-6}
    \cmidrule(lr){7-9}
    & \textbf{IoU} $\uparrow$
    & \textbf{CD} $\downarrow$
    & \textbf{CLIP} $\uparrow$
    & \textbf{FD} $\downarrow$
    & \textbf{KD} $\downarrow$
    & \textbf{CLIP} $\uparrow$
    & \textbf{FD} $\downarrow$
    & \textbf{KD} $\downarrow$ \\
    \midrule

    Single Codebook
    & 0.832 & 0.017
    & 38.27 & 18.43 & 0.13
    & 87.96 & 11.74 & 0.09 \\

    \rowcolor{gray!8}
    \textbf{\modelnamegradient}
    & \best{0.864} & \best{0.013}
    & \best{39.01} & \best{16.50} & \best{0.10}
    & \best{89.21} & \best{9.23} & \best{0.06} \\
    \bottomrule
  \end{tabular}%
  }
  \vspace{-0.2cm}
\end{table*}

\noindent\textbf{Scale-Specific Codebooks.}
We ablate whether each octree scale requires its own VQ codebook by replacing the scale-specific codebooks $\{\mathcal{C}^{(s)}\}_{s=1}^{S}$ with a single shared codebook of equivalent total capacity. Sharing degrades the tokenizer itself
(IoU drops from 0.864 to 0.832, CD rises from 0.013 to 0.017), and the
loss propagates cleanly to every downstream task. Text-to-3D CLIP falls
by 0.74 and FD rises by 1.93, while Image-to-3D CLIP falls by 1.25 and
FD rises by 2.51. These results indicate that scale-specific codebooks allow each vocabulary to specialize to the geometric primitives of its resolution, providing a cleaner basis for the scale-routing head to make coarse-to-fine anchor decisions.\looseness-1

\section{Additional Qualitative Results}\label{app:additional_qualitative}
\begin{figure}[t!]
    \centering
    \includegraphics[width=0.99\linewidth]{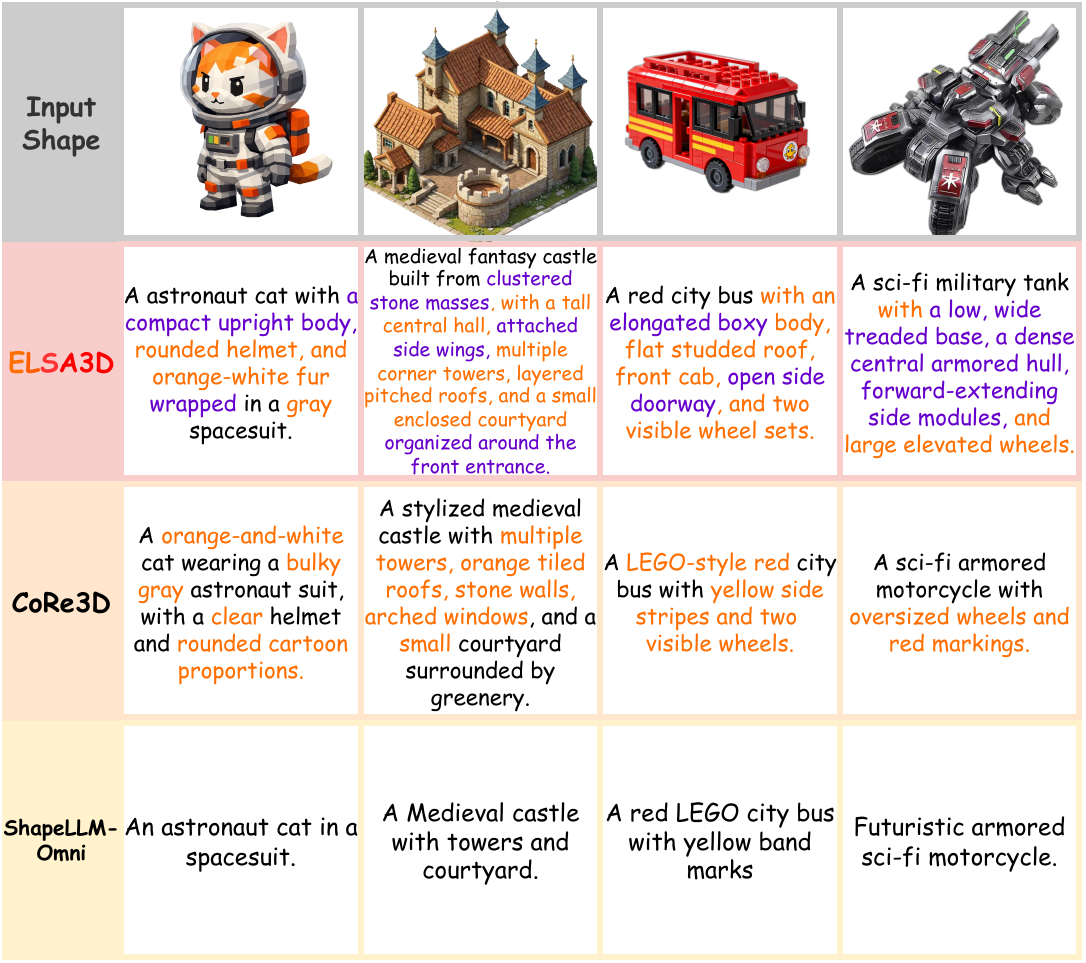}
    \vspace{-0.3cm}
    \caption{\textbf{Qualitative 3D Object Captioning Comparison.}
    \modelname{} produces more detailed and geometrically grounded captions than prior unified 3D models. \textcolor{orange}{Orange} and {purple} text highlight fine-grained structural and appearance details captured beyond the ShapeLLM-Omni baseline.}
    \label{fig:app_qual_3d_captioning}
\end{figure}

\noindent\textbf{Qualitative 3D Captioning.}
Figure~\ref{fig:app_qual_3d_captioning} provides qualitative examples of 3D object captioning.
Given the same input shape, ShapeLLM-Omni produces short
category-level descriptions (``An astronaut cat in a spacesuit'').
CoRe3D adds attribute detail but stops at coarse description, whereas \modelname produces more compositional captions that describe object identity, part structure, and local appearance, such as clustered stone masses, attached side wings, layered pitched roofs, orange-white fur, and flat studded surfaces. 
These examples demonstrate that scale-aware semantic anchoring helps the model ground language in both global shape and fine geometric detail.

\begin{figure}[t!]
    \centering
    \includegraphics[width=0.99\linewidth]{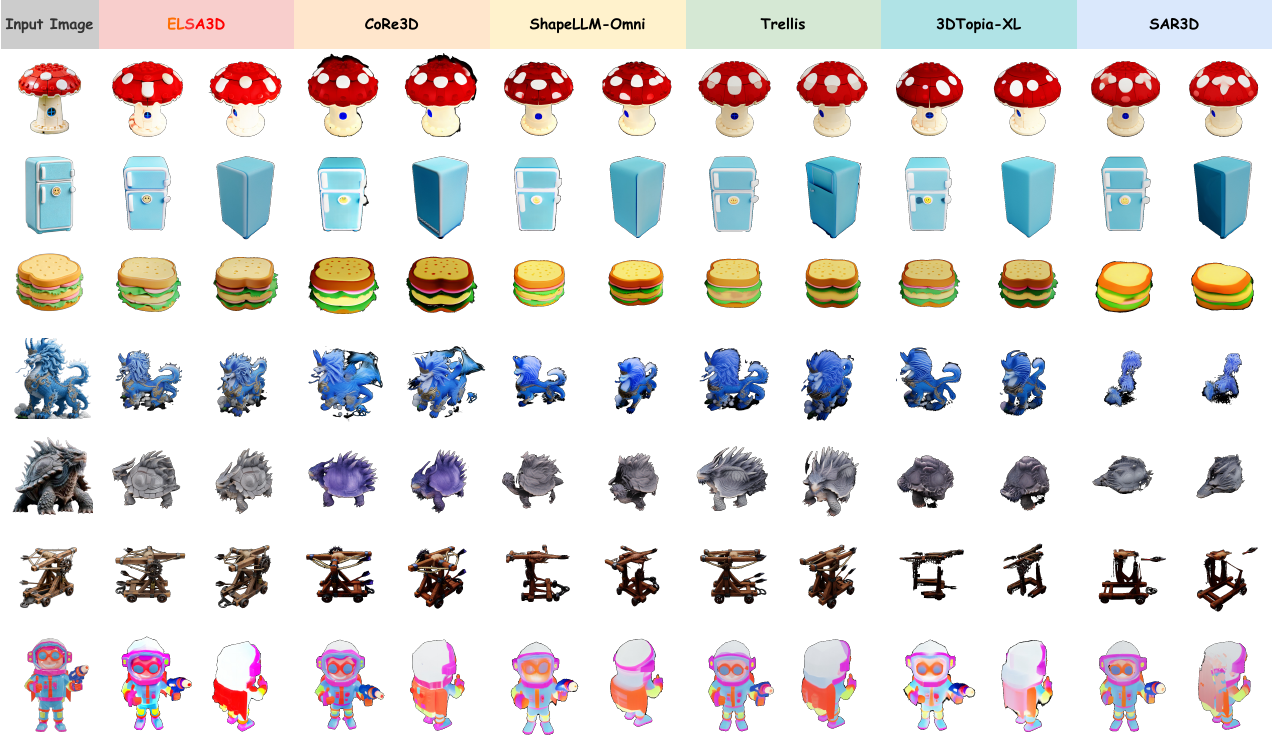}
    \vspace{-0.2cm}
    \caption{\textbf{Qualitative Image-to-3D Comparison.} Each method is shown from two rendered views. \modelname{} better preserves both global shape and local appearance cues from the input image, including thin structures, part layout, and distinctive textures.}
    \label{fig:app_qual_image_to_3d}
\end{figure}

\noindent\textbf{Additional Image-to-3D Comparisons.}
Figure~\ref{fig:app_qual_image_to_3d} provides the full image-to-3D qualitative comparison.
The examples span diverse object categories with different levels of geometric complexity, including smooth objects, thin structures, articulated parts, stylized shapes, and distinctive surface patterns.
Across these cases, \modelname{} more consistently preserves global silhouette, part layout, and local appearance cues from the input image.

\begin{figure}[t!]
    \centering
    \includegraphics[width=0.99\linewidth]{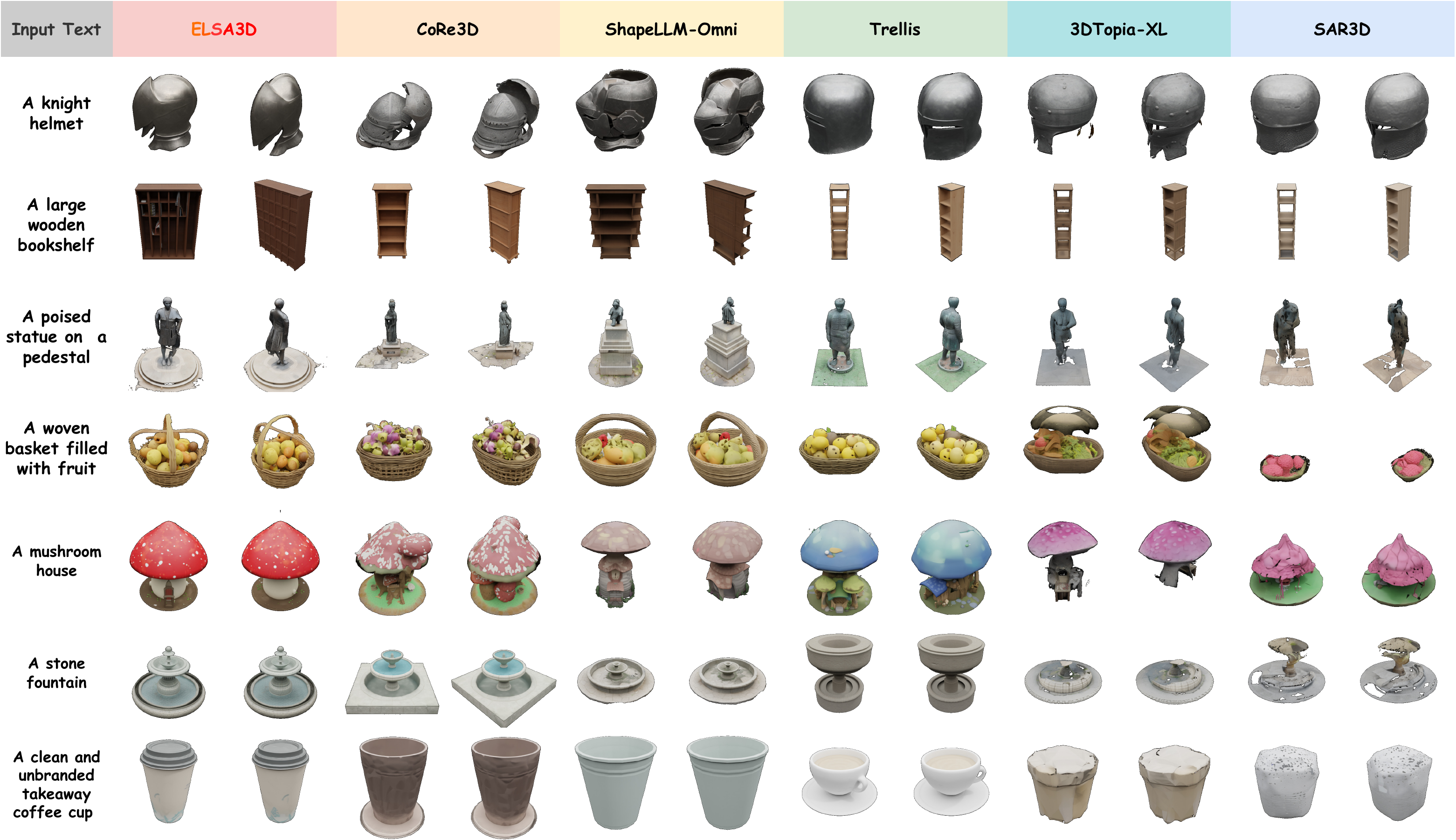}
    \vspace{-0.25cm}
    \caption{\textbf{Qualitative Text-to-3D Comparison.} \modelname{} generates objects that better satisfy both category-level intent and fine-grained prompt constraints, such as object parts, support structures, material cues, and surface appearance.}
    \label{fig:app_qual_text_to_3d}
\end{figure}

\noindent\textbf{Additional Text-to-3D Comparisons.}
Figure~\ref{fig:app_qual_text_to_3d} provides the full text-to-3D qualitative comparison.
The prompts cover category-level generation, part-level structure, support relations, material cues, and appearance constraints.
\modelname{} more faithfully follows both the intended object category and fine-grained prompt details, producing objects with more coherent part layouts and fewer generic or underspecified shapes.
These examples show that scale-aware grounding is especially useful when the input is language only, and the model must infer geometry from semantic constraints.

\begin{figure}[t!]
    \centering
    \includegraphics[width=0.99\linewidth]{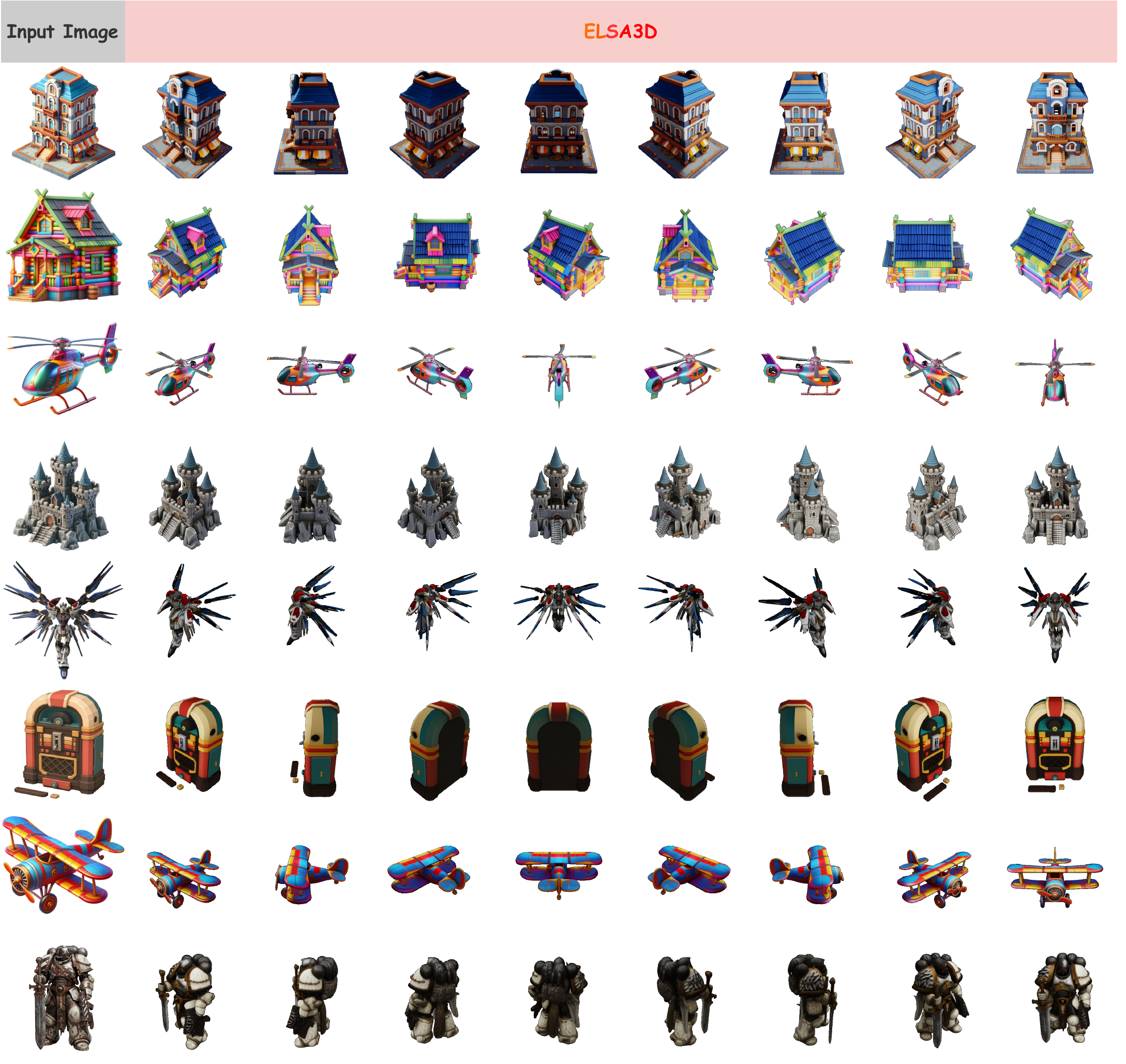}
    \vspace{-0.4cm}
    \caption{\textbf{Additional In-the-Wild Image-to-3D Results.}
    \modelname{} generates coherent 3D structure from images while preserving salient shape and appearance cues across diverse object categories.}
    \label{fig:supp_image_to_3D}
    \vspace{-0.3cm}
\end{figure}

\noindent\textbf{Additional In-the-Wild Image-to-3D Results.} \Cref{fig:supp_image_to_3D} provides additional in-the-wild image-to-3D generation results. These examples cover diverse object categories and visual appearances, including cases with complex surface details. The results further show that \modelname can recover coherent 3D structure from a single image while preserving salient shape cues and appearance patterns. This demonstrates the robustness of our unified 3D representation beyond the main quantitative benchmark.

\section{Limitations}
\label{app:limitations}
This work focuses on object-level unified 3D understanding and generation. Extending \modelname{} to large multi-object scenes, dynamic 3D content, and interactive editing remains future work. In addition, our octree tokenizer uses a fixed maximum depth, which balances quality and efficiency but may not capture all extremely fine surface details. Finally, like other generative 3D models, \modelname{} may produce plausible but incorrect geometry when prompts are ambiguous or when input images contain occluded object parts.

\section{Broader Impacts}
\label{app:broader_impact}
Unified 3D generation and understanding can support creative design, simulation, education, accessibility, robotics, and scientific visualization by making 3D content easier to generate and reason about from language or images. At the same time, such models may be misused to create deceptive or copyrighted 3D assets, generate unsafe objects, or amplify biases present in web-scale 3D and image datasets. We encourage responsible deployment with dataset documentation, provenance tracking, content filtering, and human review for high-stakes or public-facing applications. Our work focuses on object-level 3D modeling and does not claim reliability for safety-critical physical deployment without additional validation.



\end{document}